\documentclass{article}

\PassOptionsToPackage{numbers}{natbib}

\usepackage[final]{neurips_2020}

\usepackage[utf8]{inputenc} 
\usepackage[T1]{fontenc}    
\usepackage{hyperref}       
\usepackage{url}            
\usepackage{booktabs}       
\usepackage{amsfonts}       
\usepackage{nicefrac}       
\usepackage{microtype}      
\usepackage{wrapfig} 

\usepackage{graphicx}
\usepackage{pdflscape} 
\usepackage{subcaption}
\usepackage{algorithm}
\usepackage{algorithmic}
\usepackage{paralist}
\usepackage{amsmath, bm}
\usepackage{xcolor}
\usepackage{nccmath}
\usepackage{tabularx}

\usepackage{hyperref}
\usepackage[nameinlink]{cleveref} 

\usepackage{authblk}

\usepackage{xcolor}
\definecolor{mycol}{rgb}{0,0,0.65}
\hypersetup{
    colorlinks,
    linkcolor={mycol},
    citecolor={mycol},
    urlcolor={mycol}
}

\usepackage{enumitem}
\newcommand{\dollarit}[1]{$#1$}
\newcommand{\vectorize}[1]{#1}

\DeclareRobustCommand{\inlinelist}[1]{\begin{inparaenum}[(a)] #1 \end{inparaenum}}

\newcommand{\E}{\mathop{\mathbb{E}}}

\title{Advances in Black-Box VI: Normalizing Flows, Importance Weighting, and Optimization}

\author[1]{Abhinav Agrawal}
\author[1,2]{Daniel Sheldon}
\author[1]{Justin Domke}
\affil[1]{College of Information and Computer Sciences, University of Massachusetts Amherst}
\affil[2]{Department of Computer Science, Mount Holyoke College}
\affil[ ]{\texttt{\{aagrawal, sheldon, domke\}@cs.umass.edu}}

\begin{document}

\maketitle

\begin{abstract}
Recent research has seen several advances relevant to black-box variational inference (VI), but the current state of automatic posterior inference is unclear. One such advance is the use of normalizing flows to define flexible posterior densities for deep latent variable models. Another direction is the integration of Monte-Carlo methods to serve two purposes; first, to obtain tighter variational objectives for optimization, and second, to define enriched variational families through sampling. However, both flows and variational Monte-Carlo methods remain relatively unexplored for black-box VI. Moreover, on a pragmatic front, there are several optimization considerations like step-size scheme, parameter initialization, and choice of gradient estimators, for which there is no clear guidance in the literature. In this paper, we postulate that black-box VI is best addressed through a careful combination of numerous algorithmic components. We evaluate components relating to optimization, flows, and Monte-Carlo methods on a benchmark of 30 models from the Stan model library. The combination of these algorithmic components significantly advances the state-of-the-art "out of the box" variational inference.

\end{abstract}



\section{Introduction}
\label{sec:intro}

We consider the problem of automatic posterior inference. A scientist or expert creates a model $p(\vectorize{z}, x)$ for latent variables $z$ and observed variables $x$. For example, $z$ might be population-level preferences for a political candidate in each district in a country, while $x$ is an observed set of polls. Then, we wish to approximate the posterior $p(\vectorize{z}|x)$, i.e., determine what the observed data say about the latent variables under the specified model. 
The ultimate aim of automatic inference is that a domain expert can create a model and get answers without manually tinkering with inference details.

In practice, one often resorts to approximate inference. Markov chain Monte Carlo (MCMC) methods are widely applicable and are asymptotically exact, but sometimes slow. Variational inference (VI) approximates the posterior within a tractable family. This can be much faster but is not asymptotically exact. Recent developments led to ``black-box VI'' methods that, like MCMC, apply to a broad class of models \citep{ranganath2014black, hoffman2013stochastic, blei2017variational}.

However, to date, black-box VI is not widely adopted for posterior inference.  Moreover, there have been several advances that are relevant to black-box VI, but have been little evaluated in that context. One such advance is normalizing flows, which define flexible densities through a composition of invertible transformations \citep{rezende2015variational, papamakarios2019normalizing}. Surprisingly, normalizing flows have seen almost no investigation for black-box VI; instead, they have been used either to directly learn a density \citep{dinh2014nice, dinh2016density, papamakarios2017masked, kingma2018glow} or to bound the likelihood of a deep latent variable model $p_\theta(x) = \int p_\theta(\vectorize{z}, x) d\vectorize{z}$ \citep{rezende2015variational, kingma2016improved, durkan2019neural}. In both cases, there is no mechanistic model and little or no focus on posterior queries. 

Another major direction is the development of tighter variational objectives that integrate Monte Carlo techniques such as importance weighting \citep{burda2015importance, naesseth2017variational, maddison2017filtering}. These were also initially designed to support learning of deep latent variable models. However, further developments have shown that they can also be viewed as enriching the variational family \citep{bachman2015training,cremer2017reinterpreting,domke2018importance,Domke2019DivideAC}. Still, little is known about the practical performance of these techniques on real models. Further, importance weighting can be used in two ways. First, importance weighted \emph{sampling} can be used at inference time to obtain better approximate samples, regardless of how the approximate distribution was created. Second, importance weighting \emph{training} can be used during optimization to tighten an objective reflecting how well such a sampling scheme will work. There is little evidence in the literature about these.

	It is essential for many users that black-box VI be \emph{fully automatic}. This brings up numerous practical issues, such as initialization, step-size schedules, and the choice of gradient estimator. There have been few attempts to address this \citep{kucukelbir2017automatic} and several basic questions remain unanswered: Do recent techniques to enrich variational families such as normalizing flows and variational Monte Carlo methods improve accuracy on mechanistic models? Which of these is more important? Can optimization be made robust enough to work with these techniques ``out of the box'' on real models?
		
	\begin{figure*}[t]
	\begin{subfigure}{0.65\linewidth}
	  \centering
	  \includegraphics[page = 1, width=\linewidth]{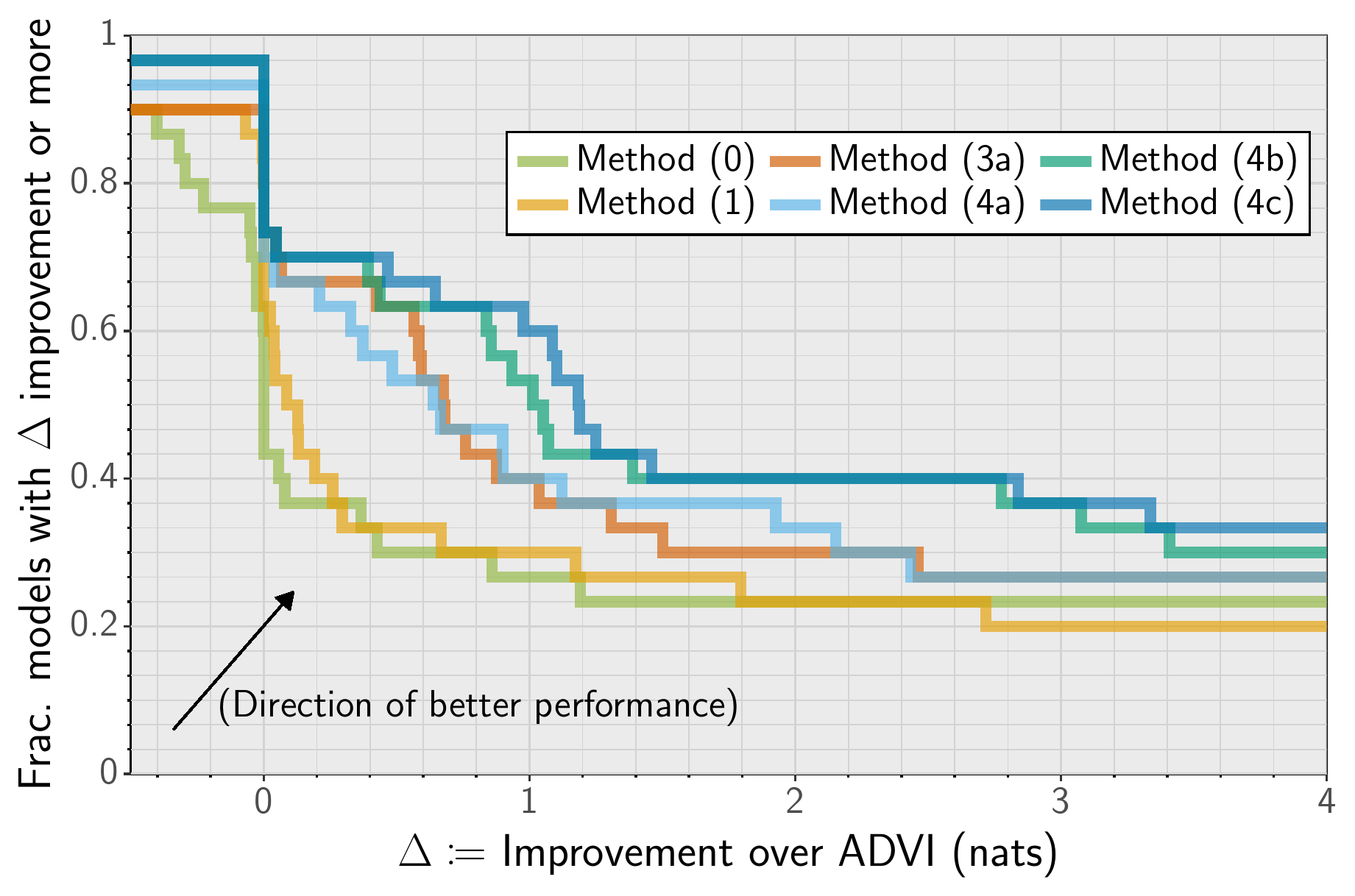}  
	\end{subfigure}
	\begin{subfigure}{0.35\linewidth}
	  \centering
	  \includegraphics[height=1.225\linewidth]{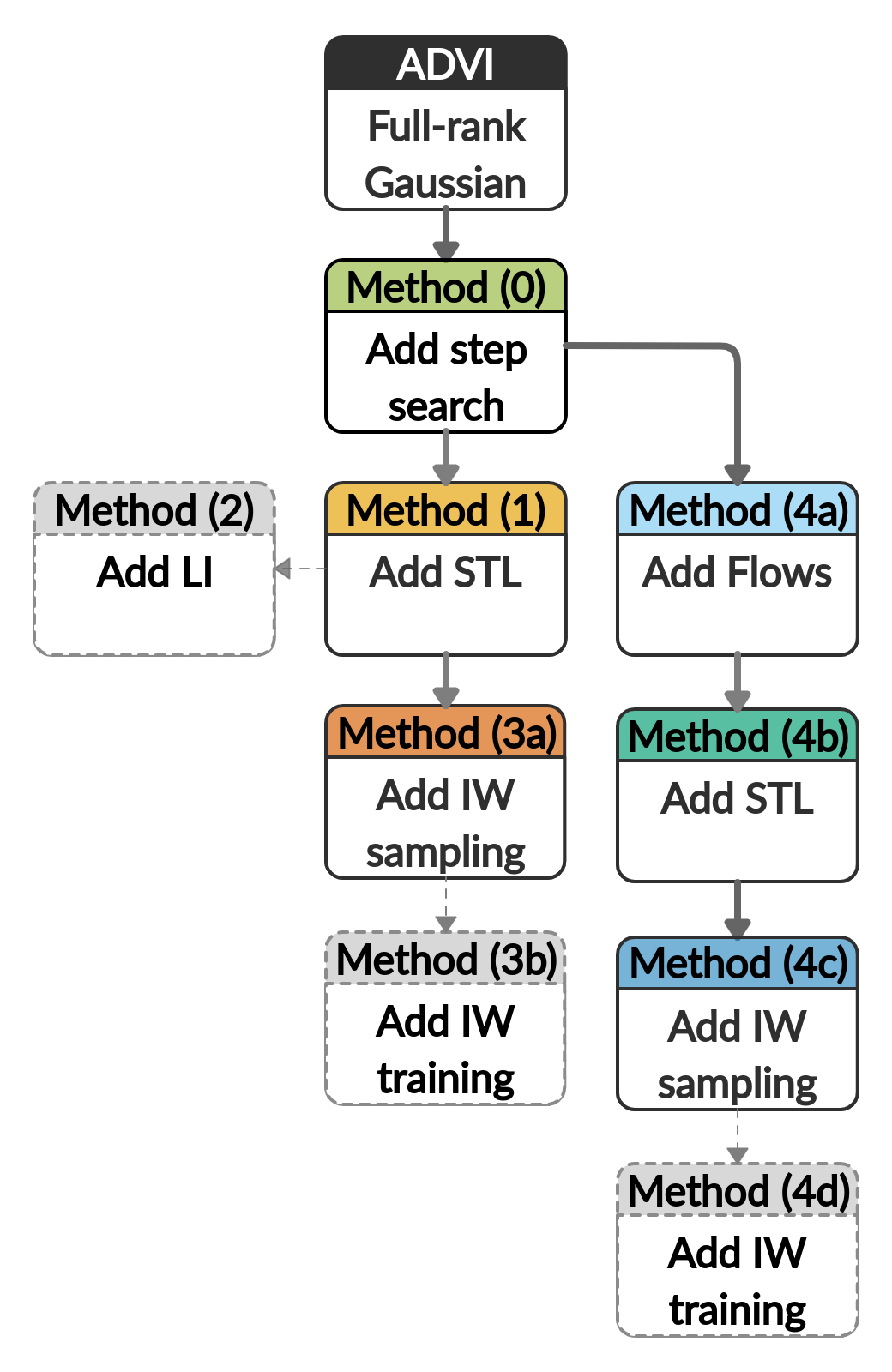}
	\end{subfigure}
	\caption{\small{\textbf{(Left) Path study}: Empirical complementary-CDF for performance improvement over ADVI. \textbf{(Right) Relationship of methods}: Solid lines show modifications that improve performance (included on the left) dotted lines show modifications with unclear benefits (not shown on the left).}}
	\label{fig:path-study-outline-combined}
	\end{figure*}

The basic hypothesis of this paper is that automatic black-box VI can be best achieved through a thoughtful integration of many different algorithmic components. We carefully measure the impact of these components on 30 real models from the Stan model library \citep{stanmodels,StanDevelopmentTeam2018}. Our primary finding is that a combination of techniques to improve optimization robustness and enrich variational families collectively yield remarkable improvements in the accuracy of black-box VI. More concretely, we make the following observations. 
\begin{itemize}[leftmargin=15pt, itemsep=3pt]

    \item A prior step-size scheme often leads to suboptimal results. Using a comprehensive search over step-sizes is far more reliable (\Cref{fig:step-size-cdf}).
	\item When using a Gaussian variational distribution, it is consistently helpful to use the sticking the landing" (STL) gradient estimator \citep{roeder2017sticking} that drops the “score term” from the evidence lower bound (ELBO) (\Cref{fig:init-cdf}).
    \item Importance-weighted sampling can be used as a post hoc method to improve the posterior found by \emph{any} inference method. This consistently improves results at minimal cost (\Cref{fig:iw_sampling,fig:nf-IW-sampling})
    \item Importance-weighted training introduces a trade-off between computational complexity and quality. The benefits are not consistent in our experiments (\Cref{fig:iw_training,fig:nf-iw-training}). 
	\item Real-NVP \citep{dinh2016density} (evaluated for the first time for black-box VI) gives excellent performance. Due to high nonlinearity, one might expect that fully-automatic optimization would be a challenge. However, when combined with our step-size scheme, it consistently delivers strong results without user-intervention (\Cref{fig:nf,fig:nf-stl,fig:nf-IW-sampling}).
	\item When using a normalizing flow variational distribution, the STL estimator is again helpful (\Cref{fig:nf-stl}). However, this uses an inverse transform, which is not efficient with all flows (\Cref{sec:nf}).
    \item The doubly-reparameterized estimator \citep{tucker2018doubly} consistently improves convergence when applied to importance weighted training (\Cref{fig:iw-training-dreg-iwae-g-nf} in appendix). However, as above, the overall value of importance-weighted training is unclear.
\end{itemize}

Our final method combines importance sampling, normalizing flows, a highly robust step-size scheme and the STL gradient estimator, to improve the performance by one nat or more on at least 60\% of the models when compared to Automatic Differentiation Variational Inference (ADVI) (see performance of Method (4c) in \Cref{fig:path-study-outline-combined}). We believe this strategy represents the state-of-the-art for fully-automatic black-box VI for posterior inference. 


\section{Problem Setup}

Directly computing the posterior is typically intractable; VI searches for the closest approximation of $p(\vectorize{z}|x)$ within a parameterized family $q_{\phi}(\vectorize{z})$ by maximizing the ELBO ~\citep{saul1996mean} 
\begin{align}
  \mathcal{L}(\phi) &= \mathbb{E}_{q_{\phi}(z)} \left[ \log p(z, x) - \log q_\phi(z) \right]  =  \log p(x) - \mathbb{KL}[q_{\phi}(z)\| p(z|x)]. \label{eq:elbo}
 \end{align}

Since KL-divergence is non-negative, $\mathcal{L}(\phi)$ lower-bounds $\log p(x)$ for all $\phi$. Moreover, when \dollarit{p} is fixed, optimizing $\mathcal{L}(\phi)$ is equivalent to minimizing the KL-divergence between $q_\phi(z)$ and $p(\vectorize{z}|x)$.

Exploiting the observations above, there are two basic uses of VI. The first is to \emph{learn} the parameters $\theta$ of a model $p_\theta(\vectorize{z},x)$ from examples of $x$. In this case, VI is used to provide a tractable objective that lower-bounds the exact likelihood. This is the basis of variational auto-encoders \citep{kingma2013auto, rezende2014stochastic} and their relatives. The second use of VI is to \emph{infer} the posterior $p(\vectorize{z}|x)$ of a fixed model. This is the setting of black-box VI \citep{ranganath2014black}, the focus of this paper.

\subsection{Rules of engagement}

This paper studies many black-box VI methods. Each method is optimized using its own objective to produce an approximate posterior $q_\phi(z)$. During optimization, all methods have the same computational budget, measured as 100 "oracle evaluations" of the $\log p$ per iteration, and are optimized for $30,000$ iterations. Then, 10,000 fresh samples are drawn from $q_\phi$ to evaluate either the ELBO in \Cref{eq:elbo} or the importance-weighted ELBO in \Cref{eq:iwelbo}, depending on the method. In either case, this gives a lower-bound on $\log p(x)$ (or equivalently an upper-bound on the KL-divergence \citep{domke2018importance}). Applying importance-weighting can be seen as an \emph{algorithmic component}, not just a "metric" since it gives a different approximation of the posterior \citep{Domke2019DivideAC}.

We evaluate each method using a benchmark of 30 models from the Stan Model library~\citep{stanmodels,StanDevelopmentTeam2018}. To remove any ambiguity, a standalone description of each method compared is given in \Cref{sec:full_method_description}.

Visualizing results across many models and inference schemes is a challenge. Simple ideas like scatterplots fail because there are many inference schemes and huge differences in $\log p(x)$ across models. Instead, we study results using "empirical complementary CDF plots". Two inference methods are compared by computing, for each model $i \in \{1,\cdots,30\}$, the difference $\Delta_i$ between the lower-bounds produced by the two methods. The idea is to plot, for each value of $\Delta$, the fraction of values $\Delta_i$ that are $\Delta$ or higher. This shows how often a given inference method improves on another by a given magnitude $\Delta$.

We start from a simple ADVI baseline model and consider changes in a single algorithmic component one by one in \Cref{fig:step-size-cdf,fig:init-cdf,fig:iw_sampling,fig:iw_training,fig:nf,fig:nf-stl,fig:nf-IW-sampling,fig:nf-iw-training}. Next, we compare the full "path" of beneficial components to ADVI in \Cref{fig:path-study-outline-combined}. Finally, we perform an ablation study removing individual algorithmic components from the final best system in \Cref{fig:ablation-study}. We conduct three independent trials for all of our experiments. For space, the path and ablation study show only a single trial in the main text, with others in the supplement. We also conduct additional experiments in \Cref{sec:extended results} including comparisons between diagonal and full-rank Gaussian VI and more comparisons of different gradient estimators. 

\section{Optimization in BBVI}
\label{sec:optimization}
Most uses of black-box VI make optimization choices on a model-specific basis. A notable exception to this is the ADVI \cite{kucukelbir2017automatic}, which we use as our baseline. 
ADVI is integrated into Stan, a state-of-the-art probabilistic programming framework for defining statistical models with latent variables \citep{carpenter2017stan}. This makes ADVI one of the most widely available black-box VI algorithms.
There are four key ideas: (1) to automatically transform the support of constrained random variables to an unconstrained space (2) to approximate the resulting unconstrained posterior with a Gaussian variational distribution, (3) to estimate the gradient of the ELBO using the reparameterization trick and (4) to do stochastic optimization with a fully-automated step-size procedure. 

 In all cases, we use the idea of transformation to unconstrained domain unchanged (1). In the rest of this section, we consider modifications to optimization ideas (3,4). In the next section, we consider generalization beyond Gaussians (2).


\label{sec:step-sizes}
	\subsection{ADVI step size search}

ADVI uses a novel decreasing step-size scheme based on a base step-size $\eta$ (we review this in \Cref{sec:advi-step-scheme}). In short, for each \dollarit{\eta \in \{0.01, 0.1, 1, 10, 100\}}, optimization is run for a small number of iterations, and the value that provides the highest final ELBO value when estimated on a fresh batch of samples is used. We use a batch of fresh 500 samples after 200 iterations. It is not clear to what degree results after a small number of iterations can give a realistic picture of how optimization will look after a large number of iterations. To test this, we propose a straightforward alternative that simply runs different steps more exhaustively.
\subsection{Comprehensive step search}
\label{sec:step-size-comp-search}
\begin{wrapfigure}{r}{0.35\textwidth}
	\centering
	\vspace{-8mm}
	\begin{small}
	¸\begin{center}
	\centerline{\includegraphics[page = 1, height=\linewidth]{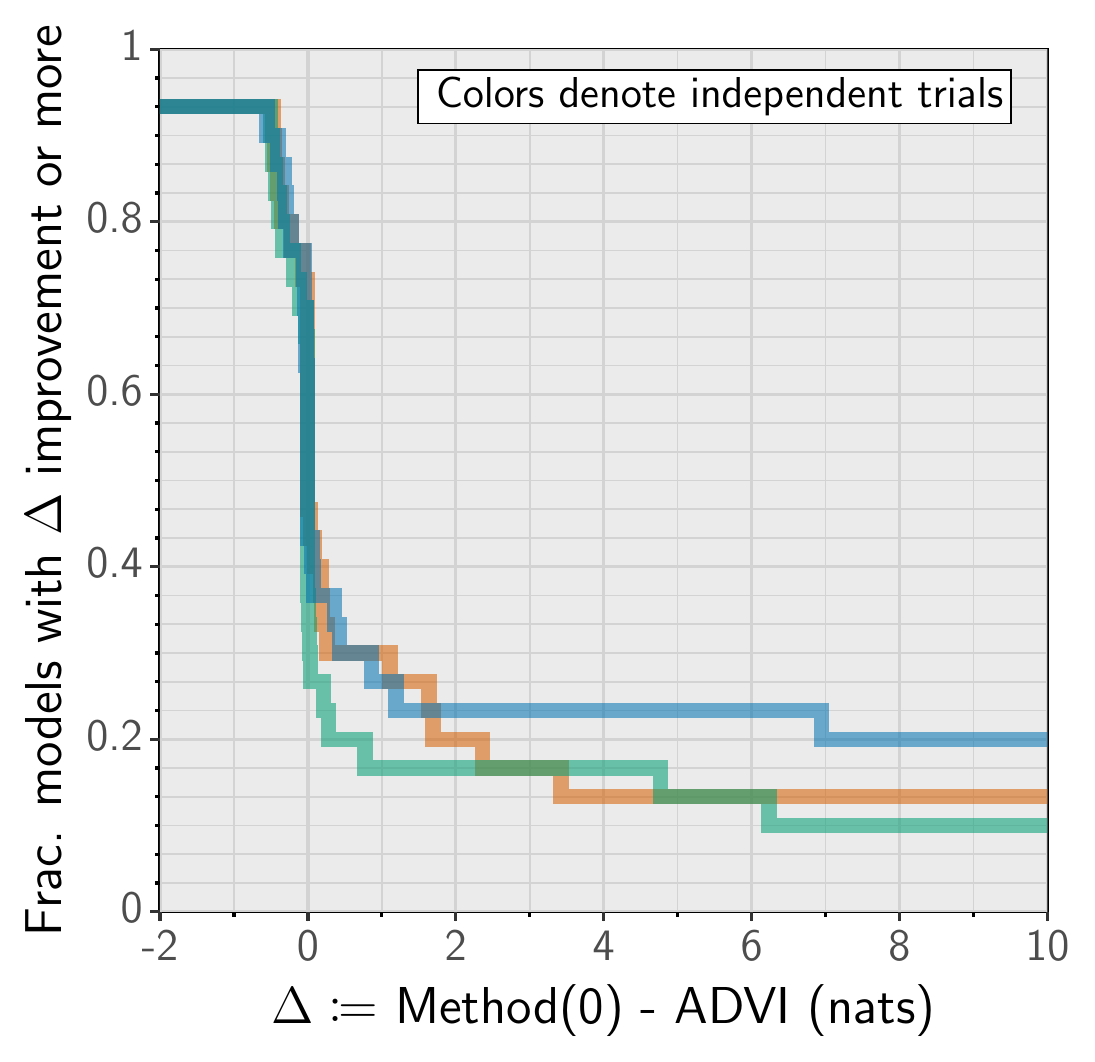}}
	\caption{\footnotesize{Using comprehensive step-size search provides significant gain of 1 nat on almost 20\% of the models (high variability is due to ADVI).}}
	\label{fig:step-size-cdf}
	\end{center}
	\end{small}
	\vspace{-8mm}
	\end{wrapfigure}
We perform optimization using Adam \citep{kingma2014adam}, and we comprehensively search for the Adam step-size \dollarit{\eta} in the range $\frac{0.1}{D}[1, {B^{-1}}, {B^{-2}}, {B^{-3}}, {B^{-4}} ],$ where $D$ is the number of latent dimensions of the model and $B$ is a decay constant; we use $B=4$. For each step size in this range, we optimize for a fixed number of iterations while keeping the step size constant. The number of iterations is constrained only by the sampling/computational budget of a user; we use 30,000 iterations for each step-size choice. 
Finally, we select the parameters from the step size that led to the best average-objective, averaged over the entire optimization trace. We found this to be a more robust indicator of performance than estimating the final ELBO using a smaller batch of fresh samples (see \Cref{sec:select-best-model} for discussion).

\Cref{fig:step-size-cdf} shows that the comprehensive search provides a significant improvement of 10 nats or more on a small fraction of the models (both schemes in the figure use the closed-form entropy estimator from \Cref{eq:justin-elbo-grad}).

\subsection{Gradient Estimation}

Most black-box VI methods are based on the reparameterization trick. Suppose that $q_\epsilon$ is a fixed distribution and $z_{\phi}(\epsilon)$ a function such that if $\epsilon \sim q_\epsilon$ then $z_{\phi}(\epsilon) \sim q_\phi$. Then, if $H(\phi)$ is the entropy of $q_\phi$, the gradient of \Cref{eq:elbo} can be written as

\begin{align}
    \nabla_{\phi}\mathcal{L}(\phi)  &= \E_{q_\epsilon (\epsilon)} \left [\nabla_{\phi} \log p(z_{\phi}(\epsilon), x) \right] + \nabla_{\phi} H(\phi). \label{eq:justin-elbo-grad}
\end{align}

The gradient can be estimated by drawing a single $\epsilon$ (or a minibatch). In some cases (e.g., Gaussians) $H$ is computed in closed form, so the above equation can be used unchanged; this is estimator used in ADVI. Alternatively, $\nabla H$ can also be estimated using the reparameterization trick, using either of

\begin{align}
\nabla_{\phi} H(\phi) = \E_{q_\epsilon (\epsilon)} \underbrace{\nabla_{\phi} \log q_\phi(z_{\phi}(\epsilon))}_\text{"Full" estimator}  = \E_{q_\epsilon (\epsilon)} \underbrace{\left (\nabla_{\phi} \log q_\theta(z_{\phi}(\epsilon)) \right)_{\theta=\phi}}_{\text{"STL" estimator}}. \label{eq:entropy-gradients}
\end{align}

The second option, known as STL estimator \citep{roeder2017sticking}, "holds $\phi$ constant" under the gradient. This is valid since $\E_{q_\phi(z)} \nabla_\phi \log q_\phi(z) = 0$.

\begin{wrapfigure}{r}{0.35\textwidth}
    \centering
    \vspace{-6.5mm}
    \begin{small}
    \begin{center}
    \centerline{\includegraphics[page = 15, height=\linewidth]{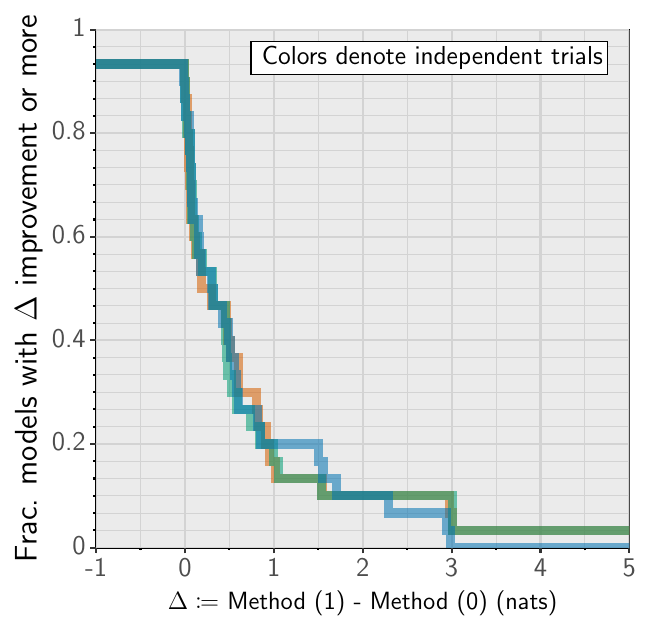}}
    \caption{\footnotesize{Adding STL to Gaussian VI is consistently helpful across the models and hurts only in minority of cases.}}
    \label{fig:init-cdf}
    \end{center}
    \end{small}
    \vspace{-16mm}
\end{wrapfigure}
\Cref{fig:init-cdf} compares the results of the STL estimator to a closed-form entropy. The STL estimator is usually preferred. We give a more comprehensive comparison of estimators in \Cref{fig:stl-vi-H estimators for g} (supplement).

\subsection{Intialization}
\label{sec:initialization}
Initialization can have a major influence on convergence. ADVI initializes to a standard Gaussian. A natural alternative for Gaussian or Elliptical variational distributions is to use Laplace's method to initialize parameters \cite{domke2018importance, Domke2019DivideAC}. 
This method uses black-box optimization to find $\hat{z}$ to maximize $\log p(z, x)$, computes the hessian of $\log p$ at $\hat{z}$ and then sets $q_\phi(z)$ to be the Gaussian whose log-density matches the local curvature of $\log p(\hat{z}, x)$. 
While Laplace's method is intuitive, it could conceivably be harmful, e.g. by providing an initialization that leads to a worse local optima. 

To the best of our knowledge, there are no existing studies of these initialization schemes. \Cref{fig:laplace-neutral} (supplement) uses the previous comprehensive step search and compares the results of adding Laplace's initialization (LI). 
A similar fraction of models are helped and harmed by the change. 

\section{Enriched Variational Families}
\label{sec:enriched}


\subsection{Monte Carlo Objectives}
\label{sec:fam-iw}

VI tries to approximate $p(z|x)$ with $q_\phi(z)$. If the approximation is inexact, Monte Carlo methods can often improve it. \citet{burda2015importance} introduced the "importance-weighted" ELBO (IW-ELBO) to use in place of the conventional ELBO when training a latent-variable model. This is
\begin{wrapfigure}{r}{0.35\textwidth}
	\centering
	\vspace{2mm}
	\begin{small}
	\begin{center}
	\centerline{\includegraphics[page = 2, height=\linewidth]{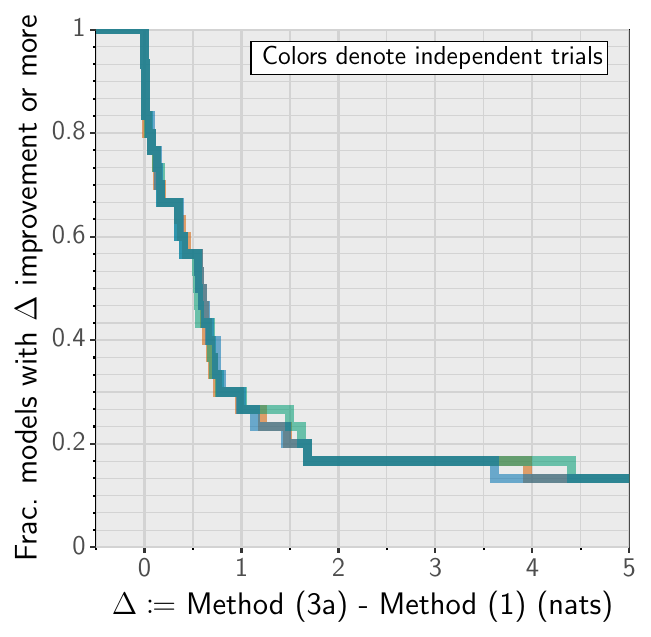}}
	\caption{\footnotesize{IW-sampling greatly improves the results of Gaussian VI.}}
	\label{fig:iw_sampling}
	\end{center}
	\end{small}
	\vspace{-12mm}
\end{wrapfigure}
\begin{align}
	\medmath{\mathcal{L}_{M}(\phi)} &= \medmath{\E_{q_\phi (z_1)\cdots q_\phi(z_M) } \left [ \log \frac{1}{M}\sum_{m=1}^{M} \frac{p(x,z_{m})}{q_\phi(z_{m})} \right] \leq \log p(x)}.\label{eq:iwelbo}
	\end{align}

Here, \dollarit{z_1  \hdots z_M} are iid samples from \dollarit{q_{\phi}(z)}. This reduces to the standard ELBO when $M=1$. The IW-ELBO increases with \dollarit{M} and approaches $\log p(x)$ asymptotically.

The IW-ELBO is a measure of the accuracy of self-normalized importance sampling on $p$ using $q_\phi$ as a proposal distribution. Define $q_{M,\phi}$ to be the distribution that results from drawing $M$ samples from $q_\phi$ and then selecting one in proportion to the self-normalized importance weights $p(z_m,x)/q(z_m)$. Then, $\mathcal{L}_M$ is a relaxation of the ELBO defined between $q_{M,\phi}$ and the target $p$ \citep{bachman2015training, cremer2017reinterpreting, naesseth2017variational}. Alternatively, $\mathcal{L}_M$ can be seen as a traditional ELBO between distributions that {\em augment} $q_{M,\phi}$ and $p$ \cite{domke2018importance}.

For posterior approximation, importance weighting can be applied in two ways.
\begin{enumerate}[leftmargin=15pt, itemsep=3pt]
	\item {\bf Importance-weighted sampling}. For {\em any} distribution, importance-weighting can be applied at test time to improve the quality of the posterior approximation. We know this because $\mathcal{L}_M(\phi) \geq \mathcal{L}(\phi)$ and thus $\mathbb{KL}[q_{M,\phi}(z)\| p(z|x)] \leq \mathbb{KL}[q_{\phi}(z)\| p(z|x)]$ \citep{domke2018importance}. This should not be seen as a metric for evaluating $\phi$. Rather, it is an algorithmic component of inference, since it yields \emph{different samples} that are actually closer to the posterior \citep{Domke2019DivideAC}.
	\item {\bf Importance-weighted training}. To find the parameters $\phi$, one can optimize the IW-ELBO. The idea is that if one intends to perform importance-weighted sampling, this directly optimizes for parameters $\phi$ that will perform well at this task. That is, optimizing the IW-ELBO implicitly makes $q_{M,\phi}(z)$ close to $p(z|x)$.
\end{enumerate}

\Cref{fig:iw_sampling} shows the results of adding importance-weighted sampling to the previous model, with $M=10$. This produces a significant benefit of 1 nat or more on 30\% of models and never hurts. This also comes at a minimal computational cost since there is no modification of the training procedure. Considerations in importance-weighted training are a bit subtle and discussed next.

\subsection{Importance-weighted training}
\label{sec:iw-training}

\begin{wrapfigure}[15]{r}{0.35\textwidth}
	\centering
	\vspace{-6mm}
	\begin{small}
	\begin{center}
	\centerline{\includegraphics[page = 7, height=\linewidth]{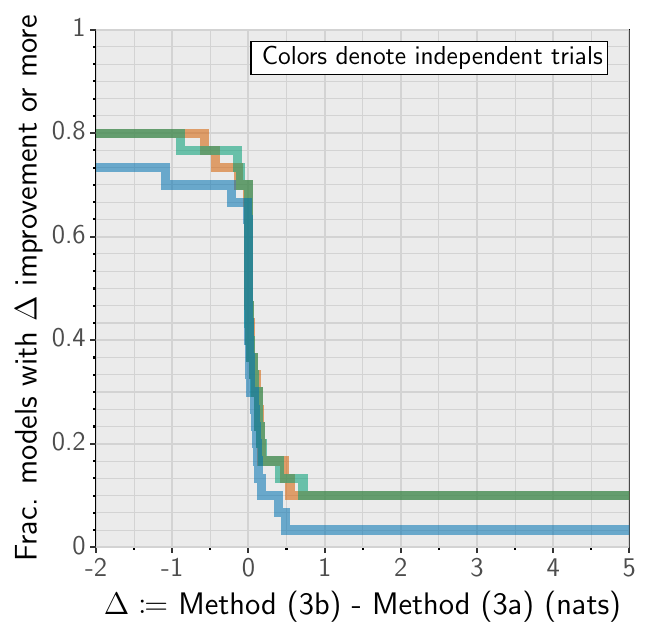}}
	\caption{\footnotesize{Adding IW-training to IW-sampling offers no clear advantage with Gaussians.}}
	\label{fig:iw_training}
	\end{center}
	\end{small}
	\vspace{-20mm}
\end{wrapfigure}
Optimizing \Cref{eq:iwelbo} requires a gradient estimator. The most obvious estimator is
\begin{align}
	\medmath{\nabla \mathcal{L}_M(\phi)} &= \medmath{\E_{q_\epsilon (\epsilon_1)\cdots q_\epsilon(\epsilon_M) } \sum_{m=1}^M \hat{w}_m \nabla_\phi \log\frac{p(z_{\phi}(\epsilon_m),x))}{q_\phi(z_{\phi}(\epsilon_m))}}  \label{eq:iw-elbo-grad},
\end{align}

where $\hat{w}_m = \frac{w_m }{\sum_{l=1}^M w_l}$ for \dollarit{w_m = \frac{p\left(x,z_{\phi}(\epsilon_m) \right)}{q_{\phi} \left( z_{\phi}(\epsilon_m)\right)}}. However, \citet{rainforth2018tighter} point out that the signal-to-noise-ratio of this gradient estimator scales as \dollarit{1/\sqrt{M}}, suggesting optimization will struggle when $M$ is large. To circumvent this problem, \citet{tucker2018doubly} introduced the "doubly reparameterized gradient estimator" (DReG), which does not suffer from the same issue. This is based on the representation of

\begin{align}
	\medmath{\nabla \mathcal{L}_M(\phi)} & = \medmath{\E_{q_\epsilon (\epsilon_1)\cdots q_\epsilon(\epsilon_M) }  \sum_{m=1}^{M} (\hat{w}_m)^{2} \nabla_\phi \log\frac{p(z_{\phi}(\epsilon_m),x))}{q_\theta(z_{\phi}(\epsilon_m))}  \Bigg\vert_{\theta=\phi}}.\label{eq:dreg-estimator}
\end{align}

Intuitively, the idea is that since $\theta$ is $\phi$ "held constant" under differentiation, the variance is reduced. The difference of the above two estimators is analogous to the difference of the two estimators of the entropy gradient in \Cref{eq:entropy-gradients}. 

We found the DReG out-performed the estimator from \Cref{eq:iw-elbo-grad}; using it added 1 nat of improvement to almost 30\% of the models (see~\Cref{fig:iw-training-dreg-iwae-g-nf}, supplement). However, the benefits are not consistent when compared to IW-sampling alone. \Cref{fig:iw_training} shows the performance {\em decays} on around 20\% of models. Thus, when keeping evaluations of $\log p$ fixed at our budget, better performance results from training the regular ELBO with the STL estimator and then using IW-sampling only.

	\subsection{Normalizing Flows}
	\label{sec:nf}

	\begin{wrapfigure}{r}{0.35\textwidth}
		\centering
		\vspace{-6.5mm}
		\begin{center}
		\centerline{\includegraphics[page = 3, width=\linewidth]{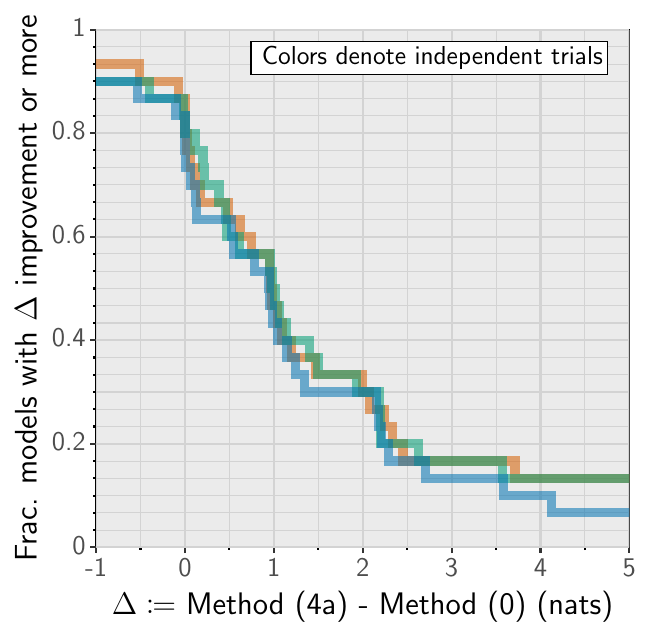}}
		\caption{\footnotesize{Substituting flows for Gaussians improves half of the models by 1 nat or more.}}
		\label{fig:nf} 
		\end{center}
		\vspace{-14mm}
	\end{wrapfigure}	
        Normalizing flows construct a flexible class of densities by applying a sequence of invertible transformations to a simple base density. \citet{rezende2015variational} first introduced normalizing flows in the specific context of approximating distributions for VI; they have since been studied in a much more general context (e.g., for density estimation \citep{dinh2016density, kingma2018glow, durkan2019neural}).

        The main idea is to transform a base density \dollarit{q_\epsilon(\epsilon)} using a diffeomorphism $T_\phi$. The variable \dollarit{z = T_\phi(\epsilon)} has the distribution 
	\begin{align}
	 q_\phi(z) &= q_\epsilon(\epsilon) |\det \nabla T_\phi(\epsilon)|^{-1} \label{eq:nf}\\
	   &= q_\epsilon(T^{-1}_\phi(z)) |\det \nabla {T^{-1}_\phi}(z)|. \label{eq:nf-inverse}
	\end{align}
	Typically, $T$ is chosen as a sequence of transforms, each of which is parameterized by a neural network. These transforms are designed so that the determinant of the Jacobian $\vert \det \nabla T_\phi(\epsilon)\vert$ can be computed efficiently. 

	We use  real-NVP--a coupling-based flow \citep{dinh2016density} with 10 successive transformations, each determined by a fully-connected network with 2 hidden layers each with 32 hidden units (see \Cref{sec:architecural-detials} for full details). \Cref{fig:nf} shows the results of using normalizing flows instead of Gaussians on the benchmark. Since there is no closed-form entropy, these results estimate it using the middle estimator from \Cref{eq:entropy-gradients}. This yields a significant improvement of 1 nat or more on around half of models.

	\begin{wrapfigure}[17]{r}{0.35\textwidth}
		\centering
		\vspace{-3mm}
		\includegraphics[page = 4, height=\linewidth]{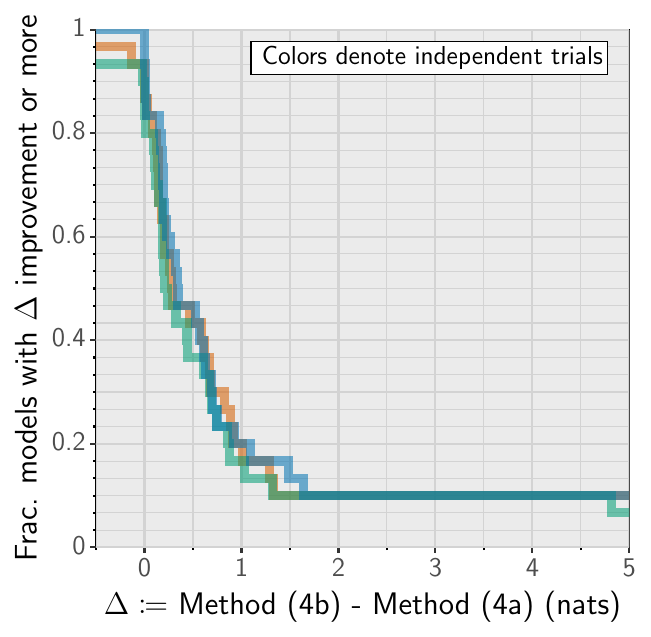}
		\caption{\footnotesize{Using the STL estimator for flows (instead of the ``full'' estimator) is typically helpful and gives huge improvements on a minority of models.}}
		\label{fig:nf-stl}
		\vspace{-8mm}
	\end{wrapfigure}
	Depending on the gradient estimator used, the inverse of $T_\phi$ may or may not be needed. This issue is somewhat subtle. In general, to evaluate the density $q_\phi$ at an arbitrary point requires computing the inverse $T^{-1}_\phi$ which may be inefficient \citep{huang2018neural, chen2019residual}. While the ELBO in \Cref{eq:elbo} requires evaluating the density $\log q_\phi$ it is only done on a point sampled from $q_\phi$. This means this can be done using only $T_\phi$--by first sampling $\epsilon$ and then transforming to $z$ there is no need to explicitly compute an inverse. Similarly, if one is to estimate  the gradient of the entropy using the "full" estimator (middle equation in \Cref{eq:entropy-gradients}), only $T_\phi$ is needed. However, the estimator that drops the score term (right equation in \Cref{eq:entropy-gradients}) is typically lower variance. The natural way to do this is to first sample from $q_\phi$, and then, in a second {\em independent} step, evaluate the sample at parameters $\theta=\phi$ that are held constant under differentiation. In this second step, the sample is treated as an arbitrary point so that derivatives will only flow to the sampling part of the algorithm. This requires the inverse $T^{-1}_\phi$.

	With  real-NVP flows, both the forward and reverse transform are efficient, so the second estimator from \Cref{eq:entropy-gradients} can be used. \Cref{fig:nf-stl} compares the results of substituting this estimator instead. In many cases, the results are much the same. However, in a significant minority of models, the new estimator yields enormous improvements.
\subsection{Importance Weighting with Normalizing Flows}
\label{sec:fam-nf-iw}

\begin{wrapfigure}{r}{0.35\textwidth}
	\centering
	\vspace{-5mm}

	\includegraphics[page = 5, height=\linewidth]{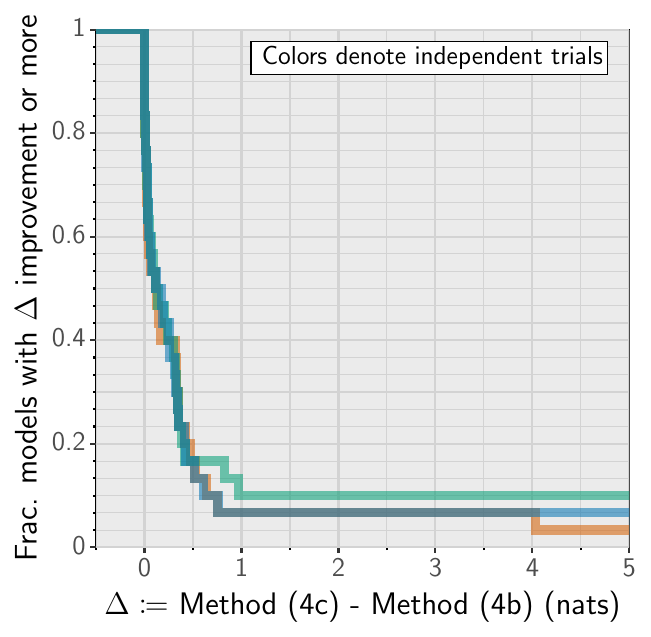}
	\caption{\footnotesize{Adding IW-sampling to normalizing flows can significantly improve some of the models and never hurts.}}
	\label{fig:nf-IW-sampling}
	\vspace{-6mm}
\end{wrapfigure}

Importance weighting and normalizing flows can be applied together. The ideas in \Cref{sec:fam-iw} are not specific to Gaussian densities. In the first experiment, we take the same optimization scheme used above, and apply importance weighted sampling only (with no change to the optimization procedure). As shown in \Cref{fig:nf-IW-sampling}, this change never hurts, provides small improvments for most models, and provides large improvements on a minority of models. Since this again comes at a minimal cost (extra work is only needed in the sampling stage) it is very worthwhile.

One can also explicitly optimize the IW-ELBO. \Cref{fig:nf-iw-training} demonstrates that adding IW-training provides large benefits on a few models but hurts performance by 1 nat or more for almost 10\% of the cases. These observations mirror the effect observed for Gaussians in \Cref{fig:iw_training}.

We now arrive at our best method: train normalizing flows with a regular ELBO objective using the lower-variance STL gradient and then add importance sampling only during inference. This is easily the go-to black-box VI strategy when we fix a budget for $\log p$ evaluations.

\section{Experiments}

\label{sec:experiments}


\paragraph{Using Stan with auto-diff packages:}
\label{sec:stan-models}

	Our experiments are based on a set of models (described in \Cref{tab:full-model-list} in supplement) from the Stan Model library \citep{StanDevelopmentTeam2018,stanmodels}. In order to test our VI variants, we designed a simple interface to use Stan models with VI algorithms implemented in Autograd, a Python automatic differentiation library \cite{maclaurin2015autograd} (refer to~\Cref{sec:auto-diff-models} for more details)

	\paragraph{Architectures:} 
	\label{sec:experiments-architectures}
	For normalizing flow methods, we use  real-NVP flow with 10 coupling layers with a fixed base architecture for all models. 
	Complete details are present in~\Cref{sec:architecural-detials}. 

	For experiments that use a full-rank Gaussian variational family, we parameterize the mean $\mu$ and the Cholesky factor $L$ of the covariance matrix $\Sigma$, such that $\Sigma = L L^{\top}$. Parameters are initialized to a standard normal when LI is not used. 

	\paragraph{Laplace's Initialization:} We use SciPy's BFGS optimize routine to maximize $\log p(z,x)$ over $z$ to get $\hat{z}$; we optimize for 2000 iterations (with default settings for other hyper-parameters) \citep{scipy}. At $\hat{z}$,  we compute the Hessian matrix $H$ using two-point finite differences with \dollarit{\epsilon = 10^{-6}}. To use LI, we set $\mu = \hat{z}$, and set $L$ such that  $LL^{\top} = (-H)^{-1}$.

	\begin{wrapfigure}[15]{r}{0.35\textwidth}
	\centering
	\vspace{-4mm}

	\includegraphics[page = 8, height=\linewidth]{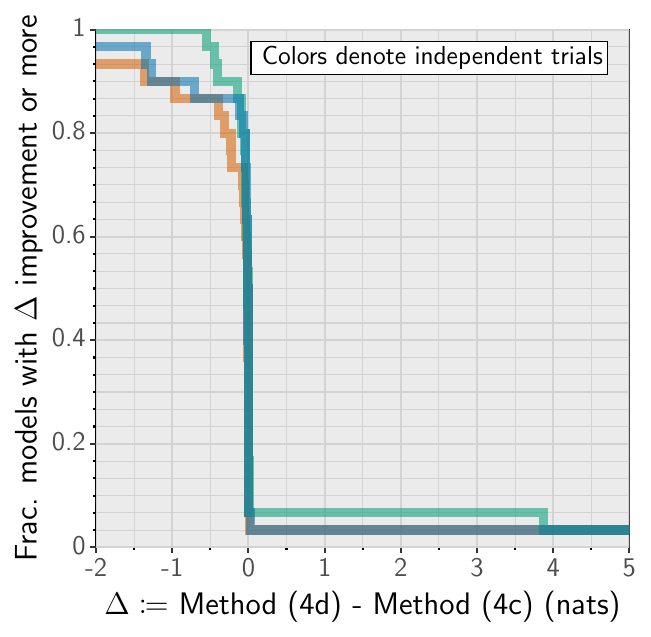}
	\caption{\footnotesize{Adding IW-training to IW-sampling offers little advantage with flows.}}
	\label{fig:nf-iw-training}
	\vspace{4mm}
	\end{wrapfigure}
	
	\paragraph{Training procedure:}\label{sec:training-procedure}

	To maintain a fair ``oracle complexity" in terms of evaluations of $\log p$, each gradient estimate averages over a batch of independent gradient estimates, such that there are always $100$ total evaluations of $\log p$ in each iteration. 
	To use \Cref{fig:nf-iw-training} as an example, IW-training method averages 10 different copies of the DReG estimator at each iteration (M = 10), and IW-sampling method averages 100 different copies of STL estimator (can be viewed as M = 1). For all experiments, we use the comprehensive step-size search scheme mentioned in~\Cref{sec:step-sizes}. Wherever importance weighting is used (sampling or training), \dollarit{M=10}
	\paragraph{Replicating ADVI:} For a fair comparison, we re-implement ADVI optimization in our own framework (refer to~\Cref{sec:advi-implementation} for more details). In our preliminary experiments found that the performance matched the PyStan version for the same hyper-parameter settings. 
	
\begin{wrapfigure}[13]{r}{0.35\textwidth}
		\centering
		\vspace{-7mm}
		\begin{small}
		¸\begin{center}
		\centerline{\includegraphics[page = 1, width=\linewidth]{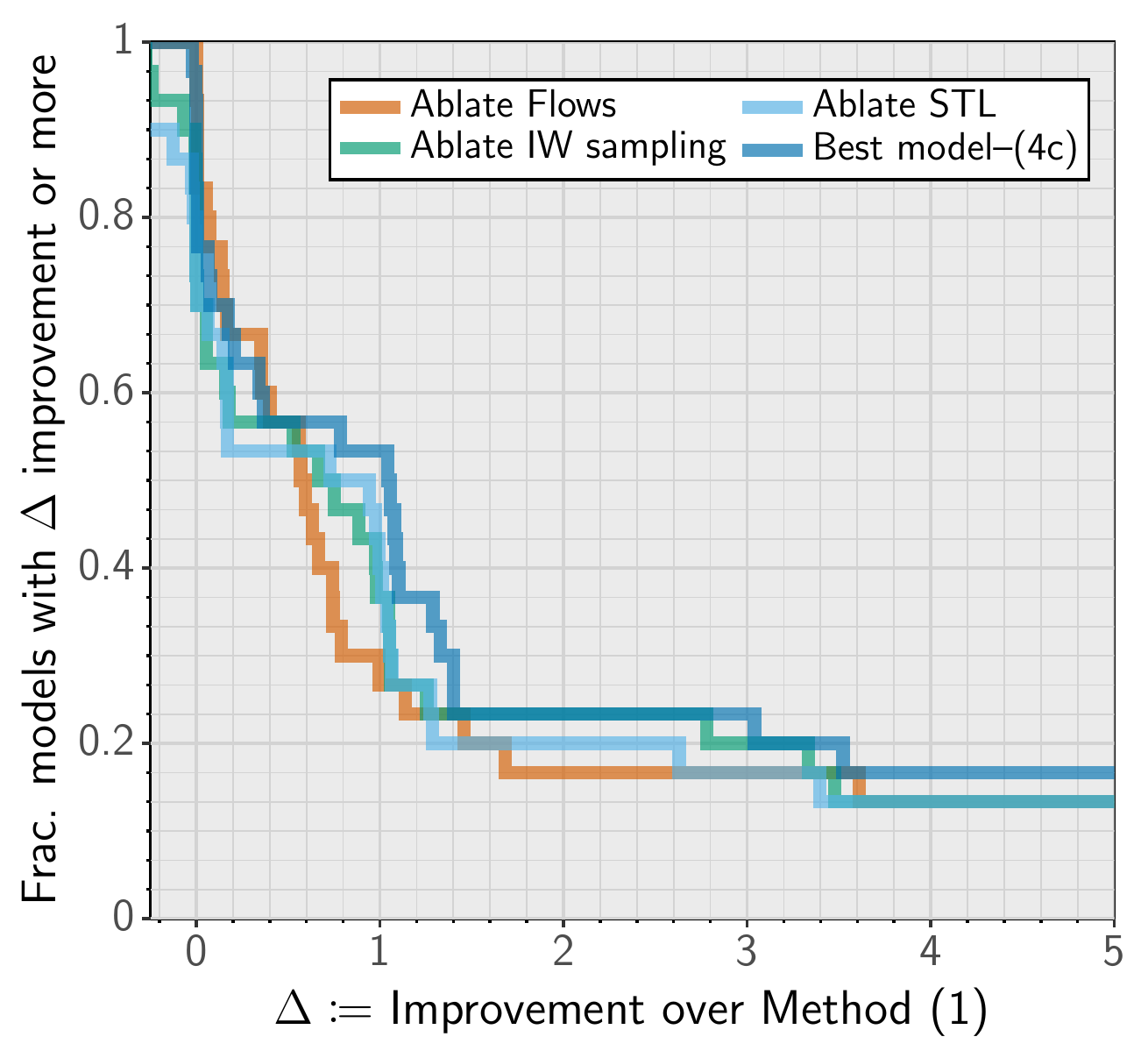}} 
		\caption{\small{\textbf{Ablation Study:} Removing flows has a strong impact; removing STL or IW-sampling has a lesser impact on the performance.}}
		\label{fig:ablation-study}
		\end{center}
		\end{small}
		\vspace{4mm}
\end{wrapfigure}

	\paragraph{Metric:}\label{sec:metric} To compare the performances, we use a fresh batch of 10,000 samples. Again, to maintain a fair ``oracle complexity'', we use 10,000 samples irrespective of the \dollarit{M} used for importance weighting. To use \Cref{fig:ablation-study} as an example, when we ablate IW-sampling, we average 10,000 copies of ELBO estimate whereas the ``Best model'' uses the 1,000 copies of the IW-ELBO estimate (M=10).
	\paragraph{Results:} 	We provide the complete tables of results with final-metric values from the independent trails in 
	\Cref{sec:complete-results}
	in the supplement. 

\subsection {Path Study}
We summarize the useful algorithmic changes in a path-study by comparing against the common ADVI baseline. 
\Cref{fig:path-study-outline-combined} presents results from independent trial (see \Cref{fig:path-study-extended} in supplement for full results). 
Across the trials, we see that with each step in the path performance improves, and the final strategy provides significant improvement of 1 nat or more on at least 50\% of the models.
\subsection{Ablation Study}
We analyze the performance of the ``best'' variational approach by taking away each of it's components individually. \Cref{fig:ablation-study} presents the ablation study for an independent trial (see \Cref{fig:ablation-study-extended} in supplement for full results).
Across the trials, each component is complementary--they add to the other and the combination performs the best.

\section{Discussion}
\paragraph{Related work.}
 \citet{pmlr-v118-fjelde20a} develop a software package (Bijector.jl) for bijective transformation of variables, that works independent of the computational framework. As a demonstration, they use normalizing flows to relax the mean-field assumption; they add coupling layers on top of a learnable diagonal Gaussian. However, the aim of our work is different--we examine the efficacy of normalizing flows and their combinations with recent inference ideas, for automating black-box VI. 

\paragraph{General comments.} We choose ELBO/IW-ELBO to compare different VI methods. Several works have reported an empirical correlation between the ELBO improvements with improvements in test-likelihoods (see \citep[Appendix, Table 3, 4, and 5]{maddison2017filtering}; \citep[Figure 4]{Chen_Tao_Zhang_Henao_Duke_2018}; \citep[Figure 4]{Mishkin_Kunstner_Nielsen_Schmidt_Khan_2018}.) and accuracy of posterior moments (see \citep[Figure 6]{domke2018importance}; \citep[Figure 7, and Figure 9 to 22]{Domke2019DivideAC}).) We expect our methods to follow similar correlations for ELBO improvements.

In our initial experiments, the performance correlated well with the number of flow layers and the number of hidden units; however, we do not aim to find the best possible flow. We use a reasonably generic real-NVP flow that allows for efficient forward and inverse; efficient inverse affords for the use of STL/DReG gradient. The use of better flow variants and optimal architectures can hopefully improve on our work.

\paragraph{Computations costs.} Trade-offs exist between the improvement in performance and the computational costs we are prepared to entertain. With more computing resources, better performance is possible.

A comprehensive step-search scheme can be done in parallel. Adding Laplace Initialization requires one to solve a maximization problem beforehand. Adding Importance Weighting training, with a fixed ``Oracle budget'' for querying $\log p$, may seem to be relatively inexpensive; however, the posterior inference scales linearly in $M$. 

Normalizing flows require more memory and computation. The parameters for full rank Gaussian scale quadratically in the number of dimensions and scale linearly for coupling-based models we consider; flows involve large constants and are expectedly slower for lower-dimensional models. Further, using the DReG or STL calculation comes at an additional cost for normalizing flows due to a second pass of $T_{\phi}^{-1}$ (see \Cref{sec:nf} for details.)

\paragraph{Conclusions.} This paper provides clear empirical evidence that systematically combining recent advances can achieve a robust inference scheme. Using step-search instead of ADVI's heuristic step selection and using normalizing flows instead of full-rank Gaussian is immensely helpful; the STL/DReG gradient consistently outperforms other available options; using Importance Weighted sampling is an almost always beneficial post-hoc step. Under the fixed sample budget, we did not find Importance Weighted training and Laplace's initialization to be consistently helpful. We hope that lessons from our work can help both researchers and practitioners.

\section*{Broader Impact}
Mechanistic modeling of natural phenomenons is a fundamental quest for modern scientists, and probabilistic models are a potent tool in such expeditions. However, automatically inferring hidden variables in these models is a challenge to date. Our approach and observations can prove instrumental for researchers in various fields--epidemiologists, ecologists, political scientists, social scientists, experimental psychologists, and others. While some other inference methods may perform better for a particular model, our ideas provide a formidable baseline.


Our framework combines several ideas--for a practitioner, this can make diagnosing the source of sub-optimality tricky. While it is possible to conduct comprehensive searches for hyper-parameters, the ability to do so hinges on access to adequate computation resources. 

\begin{ack}
  This material is based upon work supported by the National Science Foundation under Grant No. 1908577.
\end{ack}
\bibliography{bibliography}

\begin{thebibliography}{38}
\providecommand{\natexlab}[1]{#1}
\providecommand{\url}[1]{\texttt{#1}}
\expandafter\ifx\csname urlstyle\endcsname\relax
  \providecommand{\doi}[1]{doi: #1}\else
  \providecommand{\doi}{doi: \begingroup \urlstyle{rm}\Url}\fi

\bibitem[Bachman and Precup(2015)]{bachman2015training}
Philip Bachman and Doina Precup.
\newblock Training deep generative models: Variations on a theme.
\newblock In \emph{NIPS Approximate Inference Workshop}, 2015.

\bibitem[Blei et~al.(2017)Blei, Kucukelbir, and McAuliffe]{blei2017variational}
David~M Blei, Alp Kucukelbir, and Jon~D McAuliffe.
\newblock Variational inference: A review for statisticians.
\newblock \emph{Journal of the American statistical Association}, 112\penalty0
  (518):\penalty0 859--877, 2017.

\bibitem[Burda et~al.(2016)Burda, Grosse, and
  Salakhutdinov]{burda2015importance}
Yuri Burda, Roger Grosse, and Ruslan Salakhutdinov.
\newblock Importance weighted autoencoders.
\newblock In \emph{ICLR}, 2016.

\bibitem[Carpenter et~al.(2017)Carpenter, Gelman, Hoffman, Lee, Goodrich,
  Betancourt, Brubaker, Guo, Li, and Riddell]{carpenter2017stan}
Bob Carpenter, Andrew Gelman, Matthew~D Hoffman, Daniel Lee, Ben Goodrich,
  Michael Betancourt, Marcus Brubaker, Jiqiang Guo, Peter Li, and Allen
  Riddell.
\newblock Stan: A probabilistic programming language.
\newblock \emph{Journal of statistical software}, 76\penalty0 (1), 2017.

\bibitem[Chen et~al.(2018)Chen, Tao, Zhang, Henao, and
  Duke]{Chen_Tao_Zhang_Henao_Duke_2018}
Liqun Chen, Chenyang Tao, Ruiyi Zhang, Ricardo Henao, and Lawrence~Carin Duke.
\newblock Variational inference and model selection with generalized evidence
  bounds.
\newblock In \emph{International Conference on Machine Learning}, 2018.

\bibitem[Chen et~al.(2019)Chen, Behrmann, Duvenaud, and
  Jacobsen]{chen2019residual}
Tian~Qi Chen, Jens Behrmann, David~K Duvenaud, and J{\"o}rn-Henrik Jacobsen.
\newblock Residual flows for invertible generative modeling.
\newblock In \emph{NeurIPS}, 2019.

\bibitem[Cremer et~al.(2017)Cremer, Morris, and
  Duvenaud]{cremer2017reinterpreting}
Chris Cremer, Quaid Morris, and David Duvenaud.
\newblock Reinterpreting importance-weighted autoencoders.
\newblock In \emph{ICLR (Workshop)}, 2017.

\bibitem[Dinh et~al.(2015)Dinh, Krueger, and Bengio]{dinh2014nice}
Laurent Dinh, David Krueger, and Yoshua Bengio.
\newblock Nice: Non-linear independent components estimation.
\newblock In \emph{ICLR (Workshop)}, 2015.

\bibitem[Dinh et~al.(2017)Dinh, Sohl{-}Dickstein, and Bengio]{dinh2016density}
Laurent Dinh, Jascha Sohl{-}Dickstein, and Samy Bengio.
\newblock Density estimation using real {NVP}.
\newblock In \emph{ICLR}, 2017.

\bibitem[Domke and Sheldon(2018)]{domke2018importance}
Justin Domke and Daniel Sheldon.
\newblock Importance weighting and variational inference.
\newblock In \emph{NeurIPS}, 2018.

\bibitem[Domke and Sheldon(2019)]{Domke2019DivideAC}
Justin Domke and Daniel Sheldon.
\newblock Divide and couple: Using monte carlo variational objectives for
  posterior approximation.
\newblock In \emph{NeurIPS}, 2019.

\bibitem[Duchi et~al.(2011)Duchi, Hazan, and Singer]{duchi2011adaptive}
John Duchi, Elad Hazan, and Yoram Singer.
\newblock Adaptive subgradient methods for online learning and stochastic
  optimization.
\newblock \emph{Journal of machine learning research}, 12\penalty0
  (Jul):\penalty0 2121--2159, 2011.

\bibitem[Durkan et~al.(2019)Durkan, Bekasov, Murray, and
  Papamakarios]{durkan2019neural}
Conor Durkan, Artur Bekasov, Iain Murray, and George Papamakarios.
\newblock Neural spline flows.
\newblock In \emph{NeurIPS}, 2019.

\bibitem[Fjelde et~al.(2020)Fjelde, Xu, Tarek, Yalburgi, and
  Ge]{pmlr-v118-fjelde20a}
Tor~Erlend Fjelde, Kai Xu, Mohamed Tarek, Sharan Yalburgi, and Hong Ge.
\newblock Bijectors.jl: Flexible transformations for probability distributions.
\newblock In \emph{Advances in Approximate Bayesian Inference}, 2020.

\bibitem[Hoffman et~al.(2013)Hoffman, Blei, Wang, and
  Paisley]{hoffman2013stochastic}
Matthew~D. Hoffman, David~M. Blei, Chong Wang, and John Paisley.
\newblock Stochastic variational inference.
\newblock \emph{The Journal of Machine Learning Research}, 14\penalty0
  (1):\penalty0 1303--1347, 2013.

\bibitem[Huang et~al.(2018)Huang, Krueger, Lacoste, and
  Courville]{huang2018neural}
Chin-Wei Huang, David Krueger, Alexandre Lacoste, and Aaron Courville.
\newblock Neural autoregressive flows.
\newblock In \emph{ICML}, 2018.

\bibitem[Kingma and Ba(2015)]{kingma2014adam}
Diederik~P Kingma and Jimmy Ba.
\newblock Adam: A method for stochastic optimization.
\newblock In \emph{ICLR}, 2015.

\bibitem[Kingma and Welling(2014)]{kingma2013auto}
Diederik~P Kingma and Max Welling.
\newblock Auto-encoding variational bayes.
\newblock In \emph{ICLR}, 2014.

\bibitem[Kingma and Dhariwal(2018)]{kingma2018glow}
Durk~P Kingma and Prafulla Dhariwal.
\newblock Glow: Generative flow with invertible 1x1 convolutions.
\newblock In \emph{NeurIPS}, 2018.

\bibitem[Kingma et~al.(2016)Kingma, Salimans, Jozefowicz, Chen, Sutskever, and
  Welling]{kingma2016improved}
Durk~P Kingma, Tim Salimans, Rafal Jozefowicz, Xi~Chen, Ilya Sutskever, and Max
  Welling.
\newblock Improved variational inference with inverse autoregressive flow.
\newblock In \emph{NeurIPS}, 2016.

\bibitem[Kucukelbir et~al.(2017)Kucukelbir, Tran, Ranganath, Gelman, and
  Blei]{kucukelbir2017automatic}
Alp Kucukelbir, Dustin Tran, Rajesh Ranganath, Andrew Gelman, and David~M Blei.
\newblock Automatic differentiation variational inference.
\newblock \emph{The Journal of Machine Learning Research}, 18\penalty0
  (1):\penalty0 430--474, 2017.

\bibitem[Maclaurin et~al.(2015)Maclaurin, Duvenaud, and
  Adams]{maclaurin2015autograd}
Dougal Maclaurin, David Duvenaud, and Ryan~P Adams.
\newblock Autograd: Effortless gradients in numpy.
\newblock In \emph{ICML AutoML (Workshop)}, 2015.

\bibitem[Maddison et~al.(2017)Maddison, Lawson, Tucker, Heess, Norouzi, Mnih,
  Doucet, and Teh]{maddison2017filtering}
Chris~J Maddison, John Lawson, George Tucker, Nicolas Heess, Mohammad Norouzi,
  Andriy Mnih, Arnaud Doucet, and Yee Teh.
\newblock Filtering variational objectives.
\newblock In \emph{NeurIPS}, 2017.

\bibitem[Mishkin et~al.(2018)Mishkin, Kunstner, Nielsen, Schmidt, and
  Khan]{Mishkin_Kunstner_Nielsen_Schmidt_Khan_2018}
Aaron Mishkin, Frederik Kunstner, Didrik Nielsen, Mark Schmidt, and
  Mohammad~Emtiyaz Khan.
\newblock \emph{SLANG: Fast Structured Covariance Approximations for Bayesian
  Deep Learning with Natural Gradient}.
\newblock 2018.

\bibitem[Naesseth et~al.(2018)Naesseth, Linderman, Ranganath, and
  Blei]{naesseth2017variational}
Christian~A Naesseth, Scott~W Linderman, Rajesh Ranganath, and David~M Blei.
\newblock Variational sequential monte carlo.
\newblock In \emph{AISTATS}, 2018.

\bibitem[Papamakarios et~al.()Papamakarios, Nalisnick, Rezende, Mohamed, and
  Lakshminarayanan]{papamakarios2019normalizing}
George Papamakarios, Eric~T. Nalisnick, Danilo~Jimenez Rezende, Shakir Mohamed,
  and Balaji Lakshminarayanan.
\newblock Normalizing flows for probabilistic modeling and inference.
\newblock \emph{CoRR}.
\newblock URL \url{http://arxiv.org/abs/1912.02762}.

\bibitem[Papamakarios et~al.(2017)Papamakarios, Pavlakou, and
  Murray]{papamakarios2017masked}
George Papamakarios, Theo Pavlakou, and Iain Murray.
\newblock Masked autoregressive flow for density estimation.
\newblock In \emph{NeurIPS}, 2017.

\bibitem[Rainforth et~al.(2018)Rainforth, Kosiorek, Le, Maddison, Igl, Wood,
  and Teh]{rainforth2018tighter}
Tom Rainforth, Adam~R Kosiorek, Tuan~Anh Le, Chris~J Maddison, Maximilian Igl,
  Frank Wood, and Yee~Whye Teh.
\newblock Tighter variational bounds are not necessarily better.
\newblock In \emph{ICML}, 2018.

\bibitem[Ranganath et~al.(2013)Ranganath, Wang, David, and
  Xing]{ranganath2013adaptive}
Rajesh Ranganath, Chong Wang, Blei David, and Eric Xing.
\newblock An adaptive learning rate for stochastic variational inference.
\newblock In \emph{ICML}, 2013.

\bibitem[Ranganath et~al.(2014)Ranganath, Gerrish, and
  Blei]{ranganath2014black}
Rajesh Ranganath, Sean Gerrish, and David Blei.
\newblock Black box variational inference.
\newblock In \emph{Artificial Intelligence and Statistics}, 2014.

\bibitem[Rezende and Mohamed(2015)]{rezende2015variational}
Danilo~Jimenez Rezende and Shakir Mohamed.
\newblock Variational inference with normalizing flows.
\newblock In \emph{ICML}, 2015.

\bibitem[Rezende et~al.(2014)Rezende, Mohamed, and
  Wierstra]{rezende2014stochastic}
Danilo~Jimenez Rezende, Shakir Mohamed, and Daan Wierstra.
\newblock Stochastic backpropagation and approximate inference in deep
  generative models.
\newblock In \emph{ICML}, 2014.

\bibitem[Roeder et~al.(2017)Roeder, Wu, and Duvenaud]{roeder2017sticking}
Geoffrey Roeder, Yuhuai Wu, and David~K Duvenaud.
\newblock Sticking the landing: Simple, lower-variance gradient estimators for
  variational inference.
\newblock In \emph{NeurIPS}, 2017.

\bibitem[Saul et~al.(1996)Saul, Jaakkola, and Jordan]{saul1996mean}
Lawrence~K Saul, Tommi Jaakkola, and Michael~I Jordan.
\newblock Mean field theory for sigmoid belief networks.
\newblock \emph{Journal of artificial intelligence research}, 4:\penalty0
  61--76, 1996.

\bibitem[{Stan Developers}(2018)]{stanmodels}
{Stan Developers}.
\newblock \emph{Example Models}, 2018.
\newblock URL \url{https://github.com/stan-dev/example-models}.

\bibitem[{Stan Development Team}(2018)]{StanDevelopmentTeam2018}
{Stan Development Team}.
\newblock \emph{The Stan Core Library, Version 2.18.0.}, 2018.
\newblock URL \url{http://mc-stan.org}.

\bibitem[Tucker et~al.(2019)Tucker, Lawson, Gu, and Maddison]{tucker2018doubly}
George Tucker, Dieterich Lawson, Shixiang Gu, and Chris~J Maddison.
\newblock Doubly reparameterized gradient estimators for monte carlo
  objectives.
\newblock In \emph{ICLR}, 2019.

\bibitem[{Virtanen} et~al.(2020){Virtanen}, {Gommers}, {Oliphant}, {Haberland},
  {Reddy}, {Cournapeau}, {Burovski}, {Peterson}, {Weckesser}, {Bright}, {van
  der Walt}, {Brett}, {Wilson}, {Jarrod Millman}, {Mayorov}, {Nelson}, {Jones},
  {Kern}, {Larson}, {Carey}, {Polat}, {Feng}, {Moore}, {Vand erPlas},
  {Laxalde}, {Perktold}, {Cimrman}, {Henriksen}, {Quintero}, {Harris},
  {Archibald}, {Ribeiro}, {Pedregosa}, {van Mulbregt}, and
  {Contributors}]{scipy}
Pauli {Virtanen}, Ralf {Gommers}, Travis~E. {Oliphant}, Matt {Haberland}, Tyler
  {Reddy}, David {Cournapeau}, Evgeni {Burovski}, Pearu {Peterson}, Warren
  {Weckesser}, Jonathan {Bright}, St{\'e}fan~J. {van der Walt}, Matthew
  {Brett}, Joshua {Wilson}, K.~{Jarrod Millman}, Nikolay {Mayorov}, Andrew
  R.~J. {Nelson}, Eric {Jones}, Robert {Kern}, Eric {Larson}, CJ~{Carey},
  {\.I}lhan {Polat}, Yu~{Feng}, Eric~W. {Moore}, Jake {Vand erPlas}, Denis
  {Laxalde}, Josef {Perktold}, Robert {Cimrman}, Ian {Henriksen}, E.~A.
  {Quintero}, Charles~R {Harris}, Anne~M. {Archibald}, Ant{\^o}nio~H.
  {Ribeiro}, Fabian {Pedregosa}, Paul {van Mulbregt}, and SciPy 1.~0
  {Contributors}.
\newblock {SciPy 1.0: Fundamental Algorithms for Scientific Computing in
  Python}.
\newblock \emph{Nature Methods}, 17:\penalty0 261--272, 2020.
\newblock \doi{https://doi.org/10.1038/s41592-019-0686-2}.

\end{thebibliography}
\bibliographystyle{plainnat}
\newpage
\appendix{}
\section{Full Method Description}
\label{sec:full_method_description}
\subsection*{ADVI Baseline}
\begin{wrapfigure}{r}{.5\textwidth}
    \centering
    \vspace{-10mm}
    \begin{small}
    \begin{center}
    \centerline{\includegraphics[height=\linewidth]{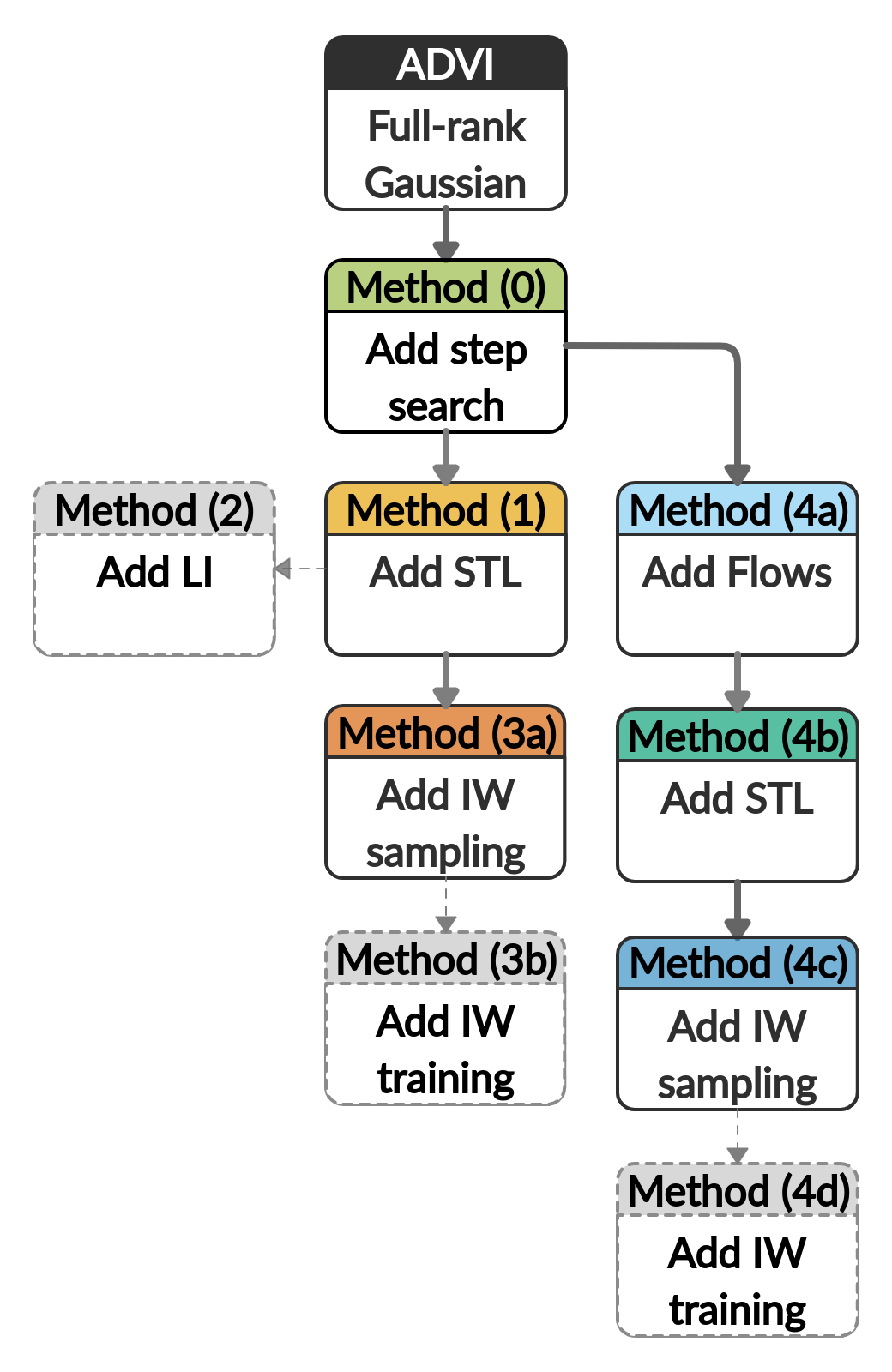}}
    \end{center}
    \end{small}
    \vspace{-22mm}
\end{wrapfigure}
Uses a full-rank Gaussian initialized to standard normal and optimizes with closed-form entropy gradient from \Cref{eq:justin-elbo-grad}; uses ADVI step-scheme for updates (see \Cref{sec:advi-implementation} for more details). Importance-weighted sampling is not used; importance-weighted training is not used (optimized standard ELBO). 
\subsection*{Method (0)}
Uses a full-rank Gaussian initialized to standard normal and optimizes with closed-form entropy gradient from \Cref{eq:justin-elbo-grad}; uses our comprehensive step-size search for updates (see \Cref{sec:step-size-comp-search} for more details). Importance-weighted sampling is not used; importance-weighted training is not used. 
\subsection*{Method (1)}
Uses a full-rank Gaussian initialized to standard normal and optimizes with the STL from \Cref{eq:entropy-gradients}; uses our comprehensive step-size search for updates. Importance-weighted sampling is not used; importance-weighted training is not used. 
\subsection*{Method (2)}
Uses a full-rank Gaussian and initializes with LI method from \Cref{sec:initialization}; optimizes with the STL from \Cref{eq:entropy-gradients} and uses our comprehensive step-size search for updates. Importance-weighted sampling is not used; importance-weighted training is not used. 
\subsection*{Method (3a)}
Uses a full-rank Gaussian initialized to standard normal and optimizes with the STL gradient from \Cref{eq:entropy-gradients}; uses our comprehensive step-size search for updates. Importance-weighted sampling is used with $M = 10$; importance-weighted training is not used. 
\subsection*{Method (3b)}
Uses a full-rank Gaussian initialized to standard normal and uses importance-weighted training with M = 10; optimizes with the DReG from \Cref{eq:dreg-estimator} and uses our comprehensive step-size search for updates. Importance-weighted training is used with M = 10 (optimizes IW-ELBO with M = 10). 
\subsection*{Method (4a)}
Uses a real-NVP normalizing flow (see \Cref{sec:architecural-detials} for architectural and initialization details) as $q_{\phi}$; optimizes with the ``full'' gradient from \Cref{eq:entropy-gradients} and uses our comprehensive step-size search for updates. Importance-weighted sampling is not used; importance-weighted training is not used.
\subsection*{Method (4b)}
Uses a real-NVP normalizing flow and optimizes with the STL gradient from \Cref{eq:entropy-gradients}; uses our comprehensive step-size search for updates. Importance-weighted sampling is not used; importance-weighted training is not used. 
\subsection*{Method (4c)}
Uses a real-NVP normalizing flow and optimizes with the STL gradient from \Cref{eq:entropy-gradients}; uses our comprehensive step-size search for updates. Importance-weighted sampling is used with M = 10; importance-weighted training is not used. 
\subsection*{Method (4d)}
Uses a real-NVP normalizing flow and uses importance-weighted training with M = 10; optimizes with  the DReG gradient from \Cref{eq:dreg-estimator} and uses our comprehensive step-size search for updates. Importance-weighted training is used with M = 10.

\section{Extended results}
\label{sec:extended results}
\subsection{Diagonal vs Full-rank Gaussian VI}

\begin{figure*}[h]
 \begin{subfigure}{0.33\linewidth}
   \centering
   \includegraphics[page = 1, width=\linewidth]{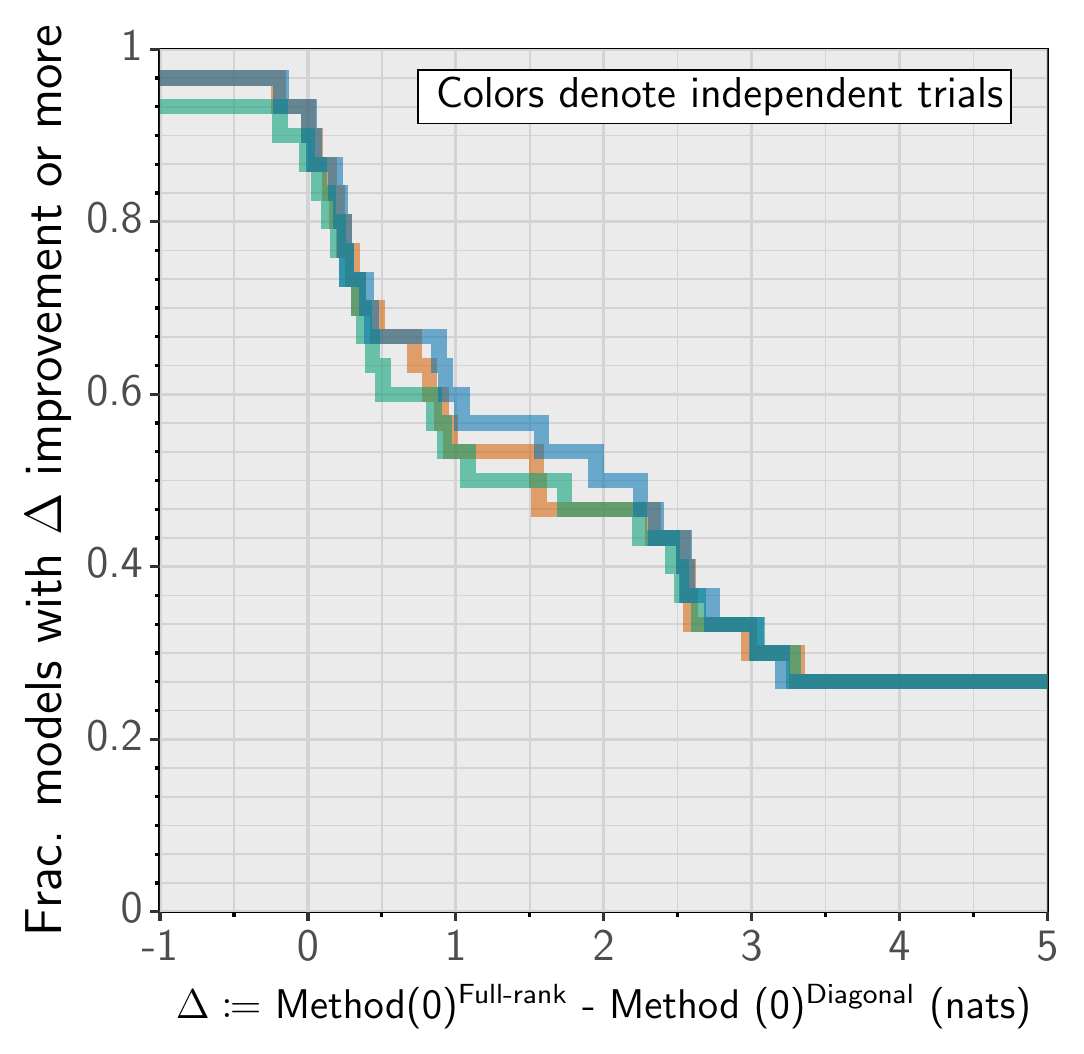}  
 \end{subfigure}
 \begin{subfigure}{0.33\linewidth}
   \centering
   \includegraphics[page = 2, width=\linewidth]{figure/diag_pairwise_comparisons.pdf}  
 \end{subfigure}
 \begin{subfigure}{0.33\linewidth}
   \centering
   \includegraphics[page = 3, width=\linewidth]{figure/diag_pairwise_comparisons.pdf}  
 \end{subfigure}
 \caption{Full-rank vs. Diagonal: \footnotesize{\inlinelist{\protect \item Method (0): closed-form entropy w/ step search, \protect \item Method (1): we replace closed-form entropy with STL gradient, and \protect \item Method (3a): we add Importance Weighted Sampling to STL gradient. In all the methods, using full-rank Gaussian improves the performance by at least 1 nats on more than half of the models. }}}
 \label{fig:diag-vs-fullrank for g}
 \end{figure*}

 In this section, we compare the performance of Gaussian VI with full-rank covariance against diagonal covariance.  
 While it is well known that full-rank covariance Gaussian distribution are more expressive, a clear experimental evidence for this is notably missing from the literature--we supplement this by experimenting with three methods from our path-study: Method (0), Method (1), and Method (3a). In \Cref{fig:diag-vs-fullrank for g}, it is easy to observe that using full-rank Gaussian improves performance by 1 nats or more on at least half of the models across the methods. When using Importance Weighted sampling--Method (3a)-- full-rank covariance Gaussians almost always improves the performance.

\subsection{Different gradient for Gaussian VI}
\begin{figure*}[h]
\begin{subfigure}{0.33\linewidth}
  \centering
  \includegraphics[page = 12, width=\linewidth]{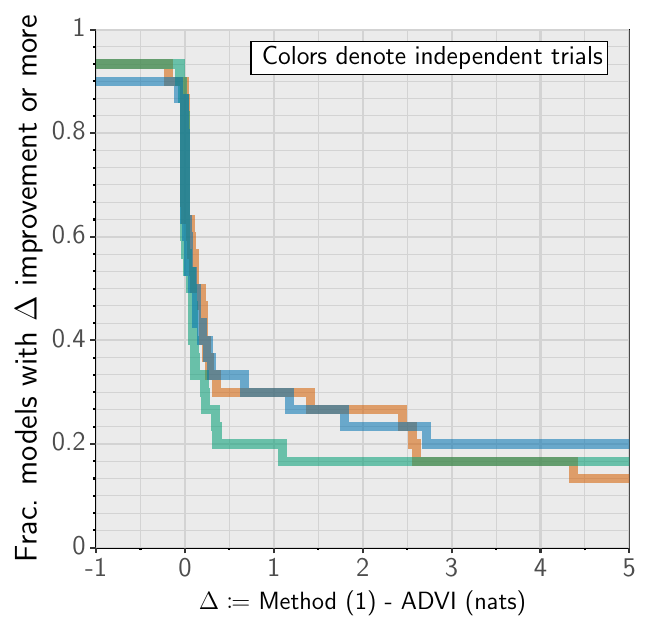}  
\end{subfigure}
\begin{subfigure}{0.33\linewidth}
  \centering
  \includegraphics[page = 14, width=\linewidth]{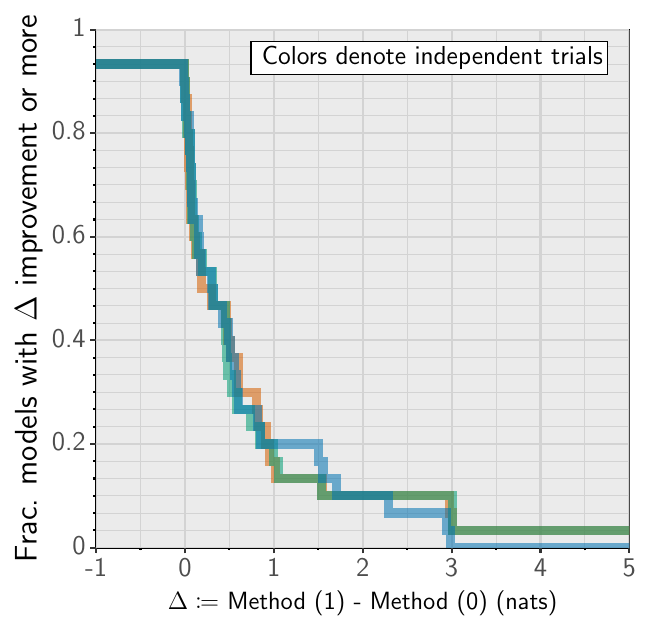}  
\end{subfigure}
\begin{subfigure}{0.33\linewidth}
  \centering
  \includegraphics[page = 13, width=\linewidth]{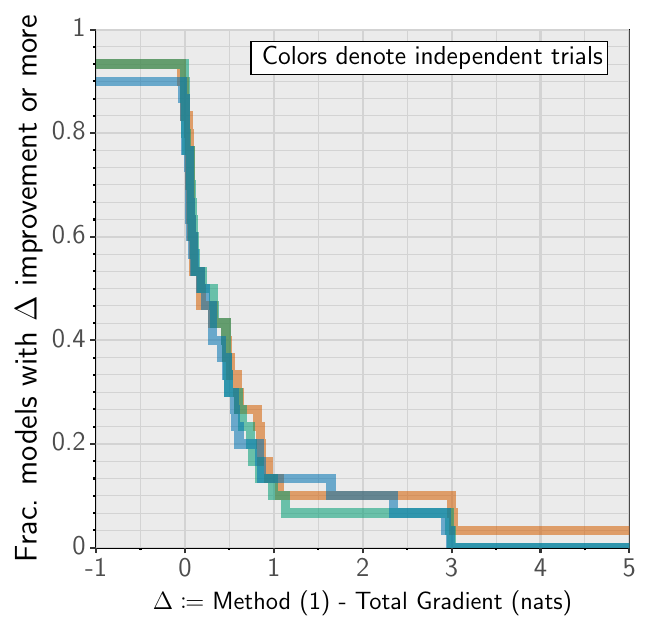}  
\end{subfigure}
\caption{\footnotesize{\inlinelist{\protect \item STL against ADVI; STL improves the performance by 1 nat or more on almost 30\% of the models. \protect \item Next, we replace the ADVI step-size scheme with our comprehensive step search. STL improves the performance on 20\% of the models by 1 nat or more. \protect \item We also compare against the middle gradient from \Cref{eq:entropy-gradients} and find that STL provides an improvement of 1 nat or more on almost 10\% of the models. In all the alternatives, STL rarely hurts.}}}
\label{fig:stl-vi-H estimators for g}
\end{figure*}

 There are three choices of gradients for the Gaussian family. First, as Gaussians have a closed-form entropy, we can use the gradient from \Cref{eq:justin-elbo-grad}; this is the gradient that ADVI uses. Second, we can alternatively use the middle gradient from \Cref{eq:entropy-gradients}. Third, we can drop the score-function term and use the STL estimator (third term in \Cref{eq:entropy-gradients}). In \Cref{fig:stl-vi-H estimators for g}, the first panel compares the performance of using STL against the ADVI implementation (ADVI step-scheme and gradient). In the second panel, we compare the performance with the closed-form estimator optimized using our comprehensive step-search. Finally, we compare against the middle gradient in \Cref{eq:entropy-gradients}. In all the alternatives, STL rarely hurts and adds significant value to several models.
\subsection{IW-training with DReG}
\begin{figure*}[h]
\begin{subfigure}{0.33\linewidth}
  \centering
  \includegraphics[page = 9, width=\linewidth]{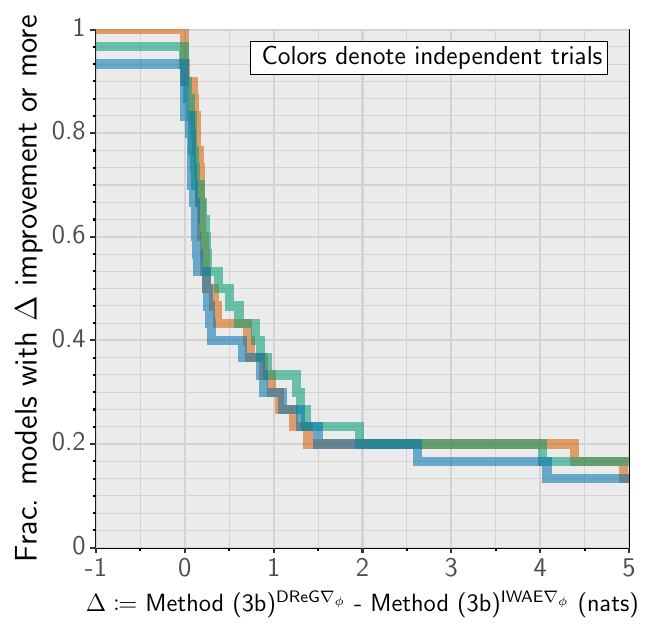}  
\end{subfigure}
\begin{subfigure}{0.33\linewidth}
  \centering
  \includegraphics[page = 10, width=\linewidth]{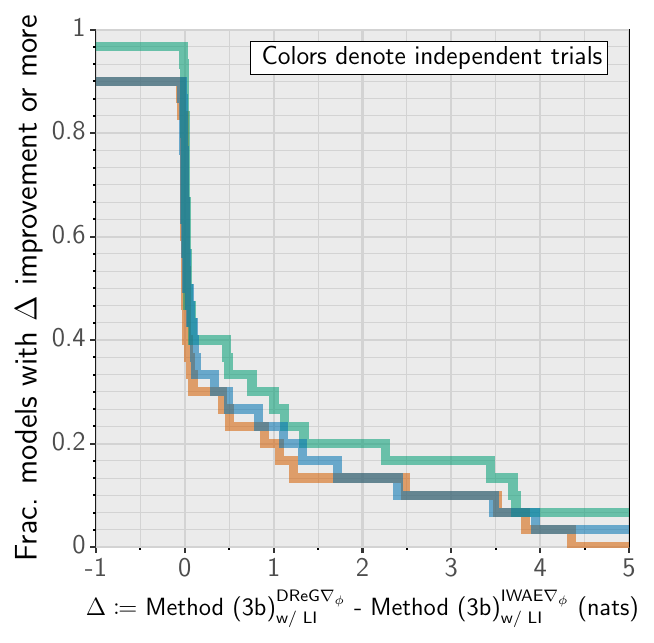}  
\end{subfigure}
\begin{subfigure}{0.33\linewidth}
  \centering
  \includegraphics[page = 11, width=\linewidth]{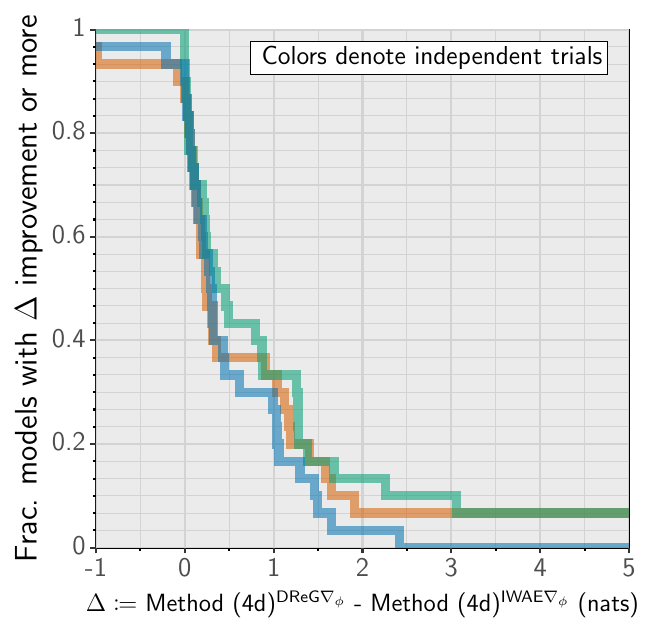}  
\end{subfigure}
\caption{\footnotesize{\inlinelist{\protect \item DReG improves the performance significantly by 1 nat on almost 30\% of the models for Gaussians when initialized with standard normal \protect \item On adding LI to the previous model, DReG adds 1 nat to around 20\% of the models \protect \item Adding DReG to IW-training of flows also helps. We observe significant improvement of 1 nat or more for almost 30\% of the models}}}

\label{fig:iw-training-dreg-iwae-g-nf}
\end{figure*}

We compare IW-training with and without DReG estimator on different possibilities and find that it consistently improved the performance. In \Cref{fig:iw-training-dreg-iwae-g-nf}, we first compare the performance with the standard IW-ELBO gradient for Gaussian families. In the second comparison, we add Laplace Initialization to both methods, IW-ELBO gradient and DReG gradient. In the final comparison, we compare DReG with regular IW-ELBO gradient for normalizing flows. Across all comparisons, DReG improves the performance and rarely hurts.

\begin{wrapfigure}{r}{0.35\textwidth}
    \centering
    \vspace{-16mm}

    \begin{small}
    \begin{center}
    \centerline{\includegraphics[page = 6, height=\linewidth]{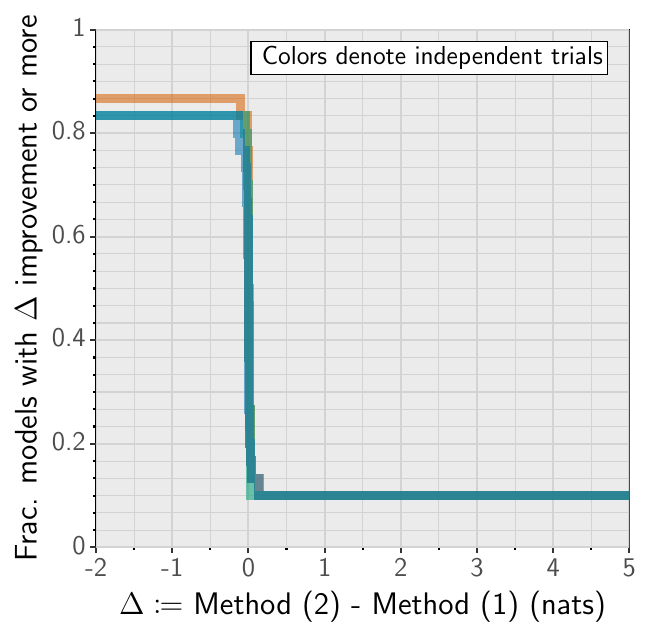}}
    \caption{\footnotesize{Adding LI to Gaussian VI is neither consistently helpful nor consistently harmful.}}
    \label{fig:laplace-neutral}
    \end{center}
    \end{small}
    \vspace{-26mm}
\end{wrapfigure}

\subsection{Path Study - full results}
\label{sec:extended-path-study}

\begin{figure*}[!ht]
\begin{subfigure}{\linewidth}
  \centering
  \includegraphics[page = 1, width=0.6\linewidth]{figure/new_figure_2_ccdf.pdf}  
\end{subfigure}
\bigskip{}
\begin{subfigure}{\linewidth}
  \centering
  \includegraphics[page = 3, width=0.6\linewidth]{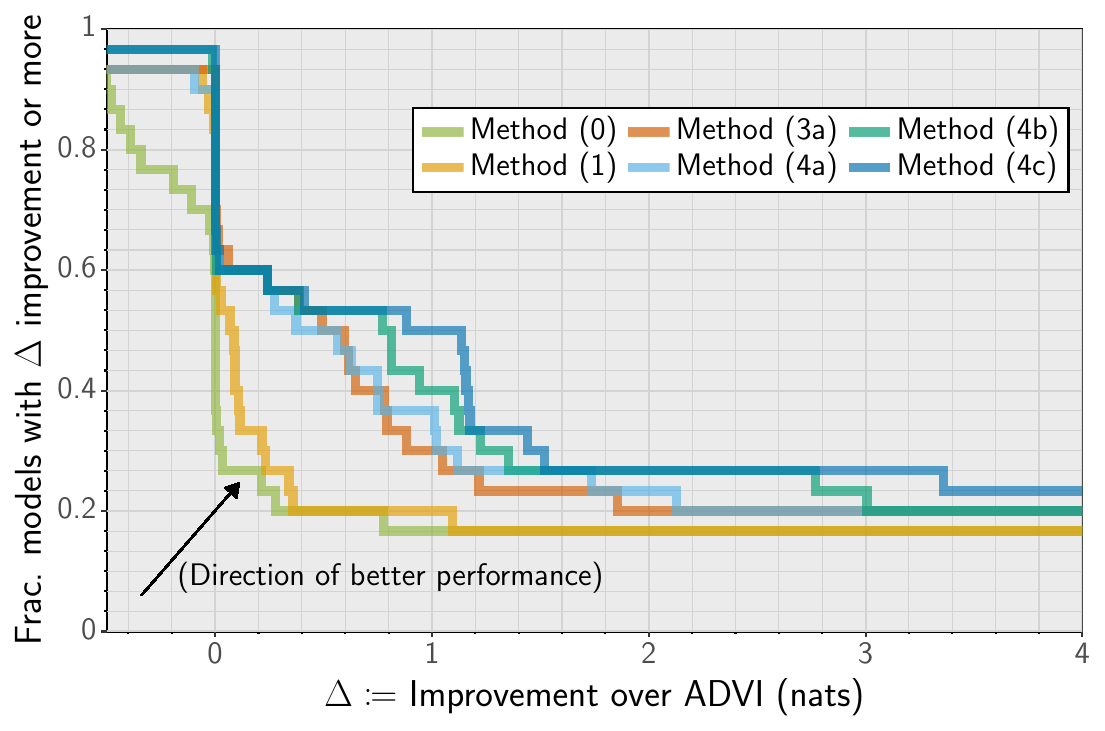}  
\end{subfigure}
\bigskip
\begin{subfigure}{\linewidth}
  \centering
  \includegraphics[page = 5, width=0.6\linewidth]{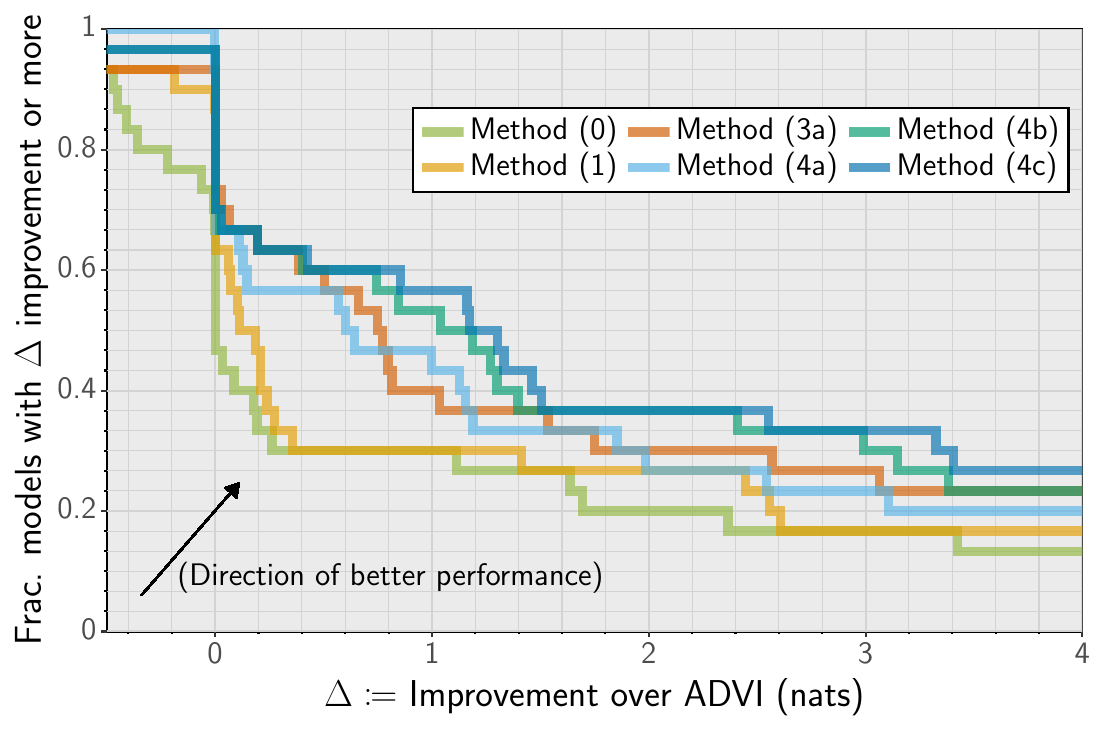}  
\end{subfigure}
\caption{\footnotesize{Across the trials: Method (1) that uses STL gradient improves over ADVI by 1 nat or more for at least 20\% of the models. Method (3a) adds the IW-sampling to (1) and improves by a nat or more on at least 30\% of the models. Method (4a) uses flow with the naive gradient estimator and achieves performance similar to (3a). Method (4b) adds the STL gradient to (4a) and improves on at least 40\% of the models by 1 nat or more. Method (4c) adds IW-sampling to (4b) and improves by 1 nat on a minimum of 50\% of the models. All our methods use comprehensive step-search and use M = 10 wherever IW-sampling is applied.}}

\label{fig:path-study-extended}
\end{figure*}

We conduct a path-study to accumulate all the useful combinations of our analysis. \Cref{fig:path-study-extended} presents the study for three independent trials. The high variation is due to the ADVI; on 10 models out of 30, ADVI diverges in at-least one trial for our implementation. If an optimization diverges, we set the improvement as zero, that is, we count the model in favor of the baseline (see \Cref{tab:advi-unstable} for values).

\subsection{Ablation Study - full results}
\label{sec:extended-ablation-study}
\begin{figure*}[!ht]
\begin{subfigure}{\linewidth}
  \centering
  \includegraphics[page = 1, width=0.6\linewidth]{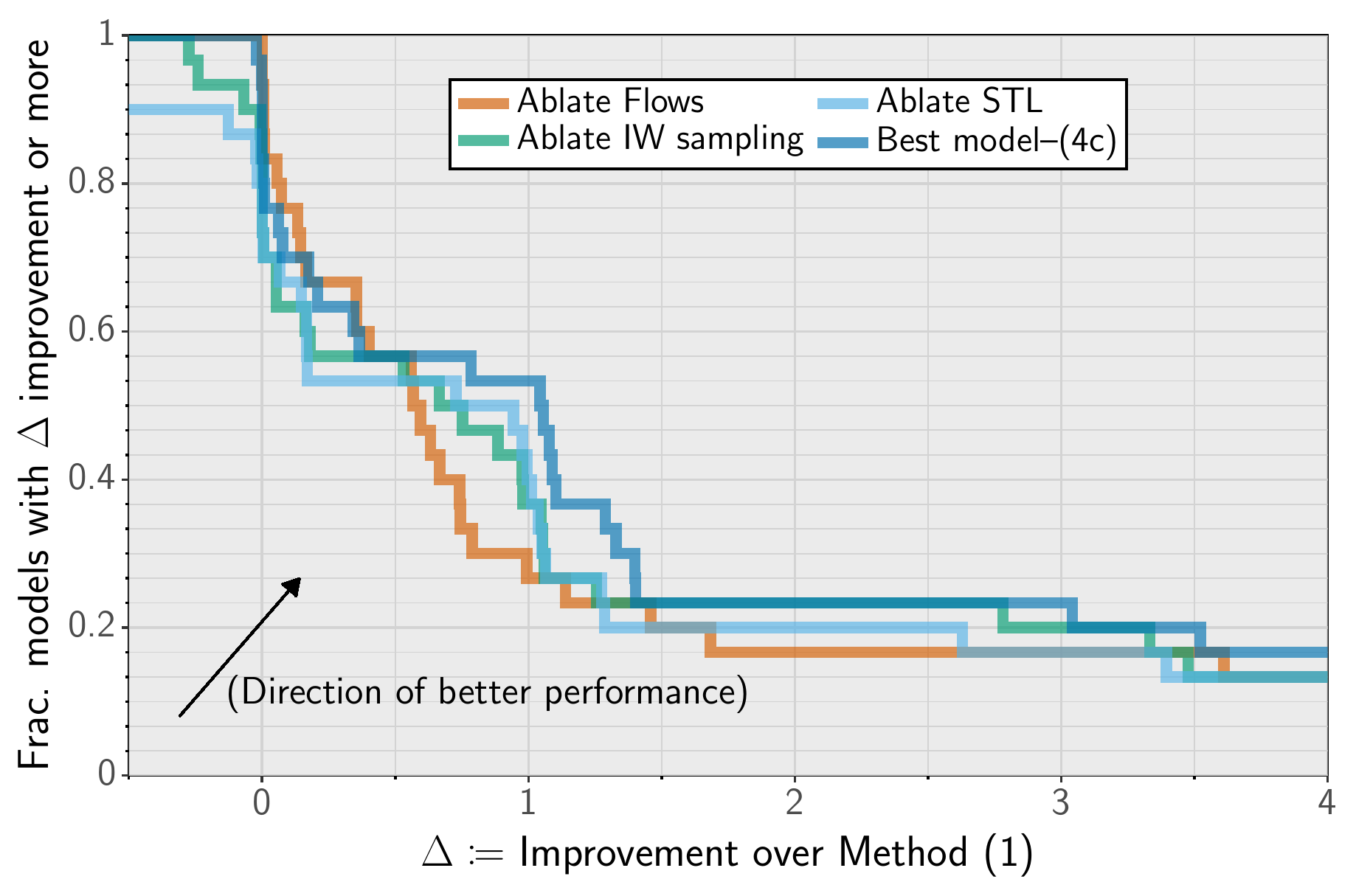}  
\end{subfigure}
\bigskip
\begin{subfigure}{\linewidth}
  \centering
  \includegraphics[page = 3, width=0.6\linewidth]{figure/new_ablation_ccdf_extended.pdf}  
\end{subfigure}
\bigskip
\begin{subfigure}{\linewidth}
  \centering
  \includegraphics[page = 5, width=0.6\linewidth]{figure/new_ablation_ccdf_extended.pdf}  
\end{subfigure}
\caption{\footnotesize{Across the trials: ablating STL observes the least decay in performance while ablating flows causes the most decrease. The effect of ablating IW-sampling lies somewhere in the middle of these two. All approaches are trained with comprehensive step-search and use M=10 wherever importance weighted sampling is used.}}

\label{fig:ablation-study-extended}
\end{figure*}

We conduct an ablation-study to analyze each component of the best performing method. \Cref{fig:ablation-study-extended} presents the study for three independent trials.

\subsection{Laplace Initialization}
\Cref{fig:laplace-neutral} compares the results of using Laplace initialization (LI) against not using it (we omitted this comparison from the main text for brevity). While there is a significant improvement on a minority of models, similar fraction observe a significant decay.  

\section{Interfacing using auto-diff packages}
\label{sec:auto-diff-models}
To interface with Stan models, we must define a new ``primitive" function in Autograd that corresponds to $\log p(x,z)$ as a function of $z$. In addtion, this also requires computing $\log p(x,z)$ itself as well as the gradient-vector product $a^\top \nabla_z \log p(z,x)$ for any vector $a$. This is easily done since PyStan interface allows access to $\log p(z,x)$ and the gradient $\nabla_z \log p(z,x)$ for any model defined in Stan. This approach has the disadvantage that high-order gradients are not possible. Similar strategies could be used with other automatic differentiation packages.

\section{ADVI}
\paragraph{Step-size scheme}
\label{sec:advi-step-scheme}
ADVI uses a novel step-size sequence inspired by adaptive step-size gradient schemes \cite{duchi2011adaptive, ranganath2013adaptive, kingma2014adam}. The update at iteration $i$ is
\begin{align}
  \phi^{(i+1)} = \phi^{(i)} - \rho^{(i)} \odot g^{(i)}, 
\end{align}
where $g^{(i)}$ is the stochastic gradient in the $i$-th iteration, $\rho^{(i)}$ is a vector of step-sizes (one per coordinate of $\phi$) and $\odot$ denotes elementwise multiplication. To determine the stepsizes, a vector $s^{(i)}$ is initialized to $s^{(1)}=(g^{(1)})^2$ and maintained recursively as
\begin{align}
  s^{(i)} = \alpha (g^{(i)})^{2}  + (1-\alpha) s^{(i-1)}, \label{eq:advi-gradient-mem}.
\end{align}
Then, the stepsizes are chosen as
\begin{align}
  \rho^{(i)}  = \frac{\eta }{ i^{1/2 + \epsilon} \times \left( \tau + \sqrt{s^{(i)}}\right)}, \label{eq:advi-eta-step-size}
\end{align}
where the square root and division are element-wise. Here $\eta>0$ is the scale of the step-size, \dollarit{i^{1/2 + \epsilon}} decays the step over time, and the $s^{(i)}$ adapts the curvature of the ELBO. $\tau=1$ and $\epsilon=10^{-16}$ are stabilizing constants.

\paragraph{Implementation details}
\label{sec:advi-implementation}
 ADVI step-scheme search for $\eta$ from \Cref{eq:advi-eta-step-size} over the range \dollarit{\{0.01, 0.1, 1, 10, 100\}} to best adapt to the size of the problem. We use 200 optimization iterations for each of these choices and then use a fresh batch of 500 samples for each step in the range to calculate final ELBO values. The step with highest final ELBO is selected as the adapted step-size; with the adapted $\eta$ we optimize for 30,0000 iterations where, at each iteration we use $100$ total \dollarit{\log p} evaluation(same as our other experiments). 

We also implement the relative-tolerance convergence criterion implemented in PyStan to detect early convergence(we use a tolerance of 0.001). Also, following the original work, we use the closed form of entropy of \dollarit{q_{\phi}} for the ADVI training objective. We make an honest attempt to the best of our abilities to re-implement the ADVI and in our preliminary experiments found that the performance matched the PyStan version for the same hyper-parameter settings. We found that the performance of ADVI was highly variable; out of the three independent trials, 9 models diverged in at least one trial. Replacing ADVI step-scheme with our comprehensive step-search saw no divergence for the closed-form entropy case that uses Adam optimizer.

\section{Implementation details for real-NVP}
\label{sec:architecural-detials}
\paragraph{Architectural details:}
We use a  real-NVP flow with 10 coupling layers for all our experiments. We define each coupling layer to be comprised of two transitions, where a single transition corresponds to affine  transformation of one part of the latent variables. For example, if the input variable for the \dollarit{k^{th}} layer is \dollarit{z^{(k)}}, then first transition is defined as     
    \begin{align}  
    z_{1:d} &= z^{(k)}_{1:d}\nonumber\\
    z_{d+1:D} &= z^{(k)}_{d+1:D} \odot \exp\big(s^{a}_{k}(z^{(k)}_{1:d})\big) + t^{a}_{k}(z^{(k)}_{1:d})).
    \label{eq:rnvp-appendix}
    \end{align}  
  where, super-script \dollarit{a} denotes first transition and sub-script \dollarit{k} denotes the \dollarit{k^{th}} layer. For the next transition, the \dollarit{z_{d+1:D}} part is kept unchanged and \dollarit{z_{1:d}} is affine transformed in a similar fashion to obtain the layer output \dollarit{z^{(k+1)}}(this time using \dollarit{s^{b}_{k} (z^{(k)}_{d+1:D}) } and \dollarit{t^{b}_{k} (z^{(k)}_{d+1:D}) }  ). This is also referred to as the alternating first half binary mask. 
  Both, scale(\dollarit{s}) and translation(\dollarit{t}) functions are parameterized by the same fully connected neural network(FNN). More specifically, for first transition in above example, a single FNN takes \dollarit{z^{(k)}_{1:d}} as input and outputs both \dollarit{s^{a}_{k}(z^{(k)}_{1:d})}  and \dollarit{t^{a}_{k}(z^{(k)}_{1:d})}. Thus, the skeleton of the FNN, in terms of the size of the layers, is as \dollarit{[d, H,H, 2(D-d)]} where, \dollarit{H} denotes the size of the two hidden layers ($H$=32 for all our experiments).  

  The hidden layers of FNN use a leaky rectified linear unit with slope = 0.01, while the output layer uses a hyperbolic tangent for \dollarit{s} and remains linear for \dollarit{t}. 

\paragraph{Parameter Initialization:}
  We initialize the parameters of the neural networks from normal distribution $\mathcal{N}(0, 0.001^{2})$. We deliberately make this choice as it corresponds to an approximate standard normal initialization for the overall normalizing flow density. To see this, first note that the output from the initialized neural networks will approximately be 0 vectors. Now, consider the affine transformation of  real-NVP: at each iteration, we scale by the exponent of $s$ and offset by $t$. Thus, the overall effect is an identity transform. As the base-distribution is fixed to a standard normal, this gives as an approximate standard normal initialization.

  \paragraph{Number of Parameters:} 
  For each transition, assuming \dollarit{d = D/2}, the parameters of the FNN can be calculated as \dollarit{ \frac{1}{2}DH + H^{2} + HD + D + 2H} where \dollarit{D} is the number latent dimensions in the model, and \dollarit{H} is the size of the two hidden layers. The first three components in the calculation corresponds to weight matrix, and the latter two take into account the bias parameters. With T coupling layers, each comprising of \dollarit{2} transitions, the overall parameter size is given by \dollarit{2T(\frac{3}{2}DH + H^{2} + D + 2H)}. We use $T$=10 and $H$=32, while $D$ depends on the problem.

\paragraph{Scaling to higher dimension models:} Real NVP based architectures scale better to higher dimensional problems as compared to Gaussians. The parameters in Gaussian scale as \dollarit{\mathcal{O}(D^{2})} while they scale linearly \dollarit{\mathcal{O}(D)} for  real-NVP, if we fix other parameters(\dollarit{T} and \dollarit{H}). However, for lower dimensional problem the number of parameters for  real-NVP is more.  

\section{Selection of Best model}
\label{sec:select-best-model}
We choose the model that achieves best average-objective, averaged over the entire optimization trace. This is different from, perhaps a more natural, final value based selection rule where one evaluates on a smaller batch of fresh samples; smaller compared to number of samples used for final metric evaluation. We found average objective to be more reliable indicator of the performance in practice. In our preliminary experiments, models selected from the maximum average-objective out-performed the ones selected based on the maximum final value; the comparison was based on the final metric value evaluated using a fresh batch of 10,000 samples.

\section{Full list of models}
We present the complete list of models used in our analysis~\Cref{tab:full-model-list}. The descriptions in the table have been manually extracted, see Stan-example model repository \citep{stanmodels} for more details. 
\begin{table}[!ht]
\caption{\label{tab:full-model-list} This table presents attributes of all the models from the Stan model library \citep{stanmodels,StanDevelopmentTeam2018} that have been used in this analysis. The attribute are $|z| = \text{\# of latent dimensions}, \,n = \text{\# of data points},\text{ and } r = \frac{\text{\# of latent dimensions}}{\text{\# of data points}} $}
\begin{center}
\begin{small}
\resizebox{0.8\linewidth}{!}{
\begin{tabularx}{\linewidth}{rlrrrX}
\toprule
 Id &           Model name &  $|z|$ &     $n$ &    $r$ &                                                                                                                                                           Model Description \\
\midrule
  1 &                 lsat &   1006 &     818 & 1.2298 &                                                                                                                         One-parameter Rasch model for LSAT student response \\
  2 &                   Mh &    388 &     385 & 1.0078 &                                                               Heterogeneity model for closed population size estimation from capture-recapture data with individual effects \\
  3 &         test\_simplex &      3 &      10 & 0.3000 &                                                                                                                                                           Simplex estimator \\
  4 &                endo3 &    184 &     626 & 0.2939 &                                                                                                    Conditional inference model for case-control study on endometrial cancer \\
  5 &           gp\_predict &    265 &     989 & 0.2679 &  Model for predicting out-of-sample observations by fitting hyperparams of a latent variable Gaussian process with exponentiated quadratic kernel and Gaussian likelihood \\
  6 &                  Mth &    394 &    1935 & 0.2036 &                                                                             Combined model for for closed population size estimation with both, time and individual effects \\
  7 &               oxford &    244 &    1226 & 0.1990 &                                                                                              A mixture model for the log odds ratio to analyze Oxford childhood cancer data \\
\bottomrule
\end{tabularx}
}
\end{small}
\end{center}
\end{table}

\section{Complete Table of results}
\label{sec:complete-results}
\begin{landscape}
\thispagestyle{empty} 
\begin{table}[h!]
\vspace{-50pt}
\caption{This table presents the results for ADVI baseline. ADVI runs with high variability in performance; our ADVI implementation diverges for at least 1 random trial for 10 out of 30 models. }
\label{tab:advi-unstable}
\begin{center}
\begin{small}
\resizebox{0.4\linewidth}{!}
{
	\begin{tabular}{llrrr}
\toprule
   & $q_{\phi}$ family & \multicolumn{3}{l}{Full-rank Gaussian} \\
   & Step-search scheme & \multicolumn{3}{l}{ADVI} \\
   & ${\nabla}_{\phi}$ & \multicolumn{3}{l}{Closed form entropy} \\
   & LI & \multicolumn{3}{l}{Not Used} \\
   & IWVI M$_{\text{training}}$ & \multicolumn{3}{l}{1} \\
   & IWVI M$_{\text{sampling}}$ & \multicolumn{3}{l}{1} \\
   & Method from Outline & \multicolumn{3}{l}{ADVI Baseline} \\
   & Independent Trial &             Trial 1 &     Trial 2 &     Trial 3 \\
\bottomrule
\end{tabular}

}
\end{small}
\end{center}
\end{table}

\end{landscape}

\begin{landscape}
\thispagestyle{empty} 
\begin{table}[h!]
\caption{\label{tab:g_closed_form} This table provides results for method that uses the closed-form entropy gradient with our comprehensive step-search scheme. We further provide {\em additional} the results by using Laplace Initialization scheme and using IW-sampling at inference time.}
\begin{center}
\begin{small}
\resizebox{\linewidth}{!}
{\begin{tabular}{llrrrrrrrrrrrr}
\toprule
   & $q_{\phi}$ family & \multicolumn{12}{l}{Full-rank Gaussian} \\
   & Step-search scheme & \multicolumn{12}{l}{Comprehensive step-search} \\
   & ${\nabla}_{\phi}$ & \multicolumn{12}{l}{Closed form entropy} \\
   & LI & \multicolumn{6}{l}{Not Used} & \multicolumn{6}{l}{Used} \\
   & IWVI M$_{\text{training}}$ & \multicolumn{6}{l}{1} & \multicolumn{6}{l}{1} \\
   & IWVI M$_{\text{sampling}}$ & \multicolumn{3}{l}{1} & \multicolumn{3}{l}{10} & \multicolumn{3}{l}{1} & \multicolumn{3}{l}{10} \\
   & Method from Outline & \multicolumn{3}{l}{(0)} & \multicolumn{3}{l}{Additional} & \multicolumn{3}{l}{Additional} & \multicolumn{3}{l}{Additional} \\
   & Independent Trial &                   Trial 1 &     Trial 2 &     Trial 3 &     Trial 1 &     Trial 2 &     Trial 3 &     Trial 1 &     Trial 2 &     Trial 3 &     Trial 1 &     Trial 2 &     Trial 3 \\
\bottomrule
\end{tabular}
}
\end{small}
\end{center}
\end{table}

\end{landscape}

\begin{landscape}
\thispagestyle{empty} 
\begin{table}[h!]
\caption{\label{tab:g_wo_stl} This table provides results for Gaussian VI that uses the ``full'' entropy gradient from \Cref{eq:entropy-gradients} with our comprehensive step-search scheme. We further provide {\em additional} the results when using Laplace Initialization scheme and using IW-sampling at inference time.}
\begin{center}
\begin{small}
\resizebox{\linewidth}{!}
{\begin{tabular}{llrrrrrrrrrrrr}
\toprule
   & $q_{\phi}$ family & \multicolumn{12}{l}{Full-rank Gaussian} \\
   & Step-search scheme & \multicolumn{12}{l}{Comprehensive step-search} \\
   & ${\nabla}_{\phi}$ & \multicolumn{12}{l}{Estimated without dropping score function term--full gradient} \\
   & LI & \multicolumn{6}{l}{Used} & \multicolumn{6}{l}{Not Used} \\
   & IWVI M$_{\text{training}}$ & \multicolumn{6}{l}{1} & \multicolumn{6}{l}{1} \\
   & IWVI M$_{\text{sampling}}$ & \multicolumn{3}{l}{1} & \multicolumn{3}{l}{10} & \multicolumn{3}{l}{1} & \multicolumn{3}{l}{10} \\
   & Method from Outline & \multicolumn{3}{l}{Additional} & \multicolumn{3}{l}{Additional} & \multicolumn{3}{l}{Additional} & \multicolumn{3}{l}{Additional} \\
   & Independent Trial &                   Trial 1 &     Trial 2 &     Trial 3 &     Trial 1 &     Trial 2 &     Trial 3 &     Trial 1 &     Trial 2 &     Trial 3 &     Trial 1 &     Trial 2 &     Trial 3 \\
\bottomrule
\end{tabular}
}
\end{small}
\end{center}
\end{table}

\end{landscape}

\begin{landscape}
\thispagestyle{empty} 
\begin{table}[h!]
\caption{\label{tab:g_w_stl}  This table provides results when using STL gradient from \Cref{eq:entropy-gradients} with our comprehensive step-search scheme. We further provide {\em additional} the results when using Laplace Initialization scheme and using IW-sampling at inference time.}
\begin{center}
\begin{small}
\resizebox{\linewidth}{!}
{
	\begin{tabular}{llrrrrrrrrrrrr}
\toprule
   & $q_{\phi}$ family & \multicolumn{12}{l}{Full-rank Gaussian} \\
   & Step-search scheme & \multicolumn{12}{l}{Comprehensive step-search} \\
   & ${\nabla}_{\phi}$ & \multicolumn{12}{l}{Estimate with STL} \\
   & LI & \multicolumn{6}{l}{Used} & \multicolumn{6}{l}{Not Used} \\
   & IWVI M$_{\text{training}}$ & \multicolumn{6}{l}{1} & \multicolumn{6}{l}{1} \\
   & IWVI M$_{\text{sampling}}$ & \multicolumn{3}{l}{1} & \multicolumn{3}{l}{10} & \multicolumn{3}{l}{1} & \multicolumn{3}{l}{10} \\
   & Method from Outline & \multicolumn{3}{l}{(2)} & \multicolumn{3}{l}{Additional} & \multicolumn{3}{l}{(1)} & \multicolumn{3}{l}{(3a)} \\
   & Independent Trial &                   Trial 1 &     Trial 2 &     Trial 3 &     Trial 1 &     Trial 2 &     Trial 3 &     Trial 1 &     Trial 2 &     Trial 3 &     Trial 1 &     Trial 2 &     Trial 3 \\
\bottomrule
\end{tabular}

}
\end{small}
\end{center}
\end{table}

\end{landscape}

\begin{landscape}
\thispagestyle{empty} 
\begin{table}[h!]
\caption{\label{tab:g_iwvi} This table provides results for importance-weighted training for Gaussian $q_{\phi}$ optimized with our comprehensive step-search scheme. We further provide {\em additional} the results when using Laplace Initialization scheme and using the regular IW-ELBO gradient of \Cref{eq:iw-elbo-grad}}
\begin{center}
\begin{small}
\resizebox{\linewidth}{!}
{\begin{tabular}{llrrrrrrrrrrrr}
\toprule
   & $q_{\phi}$ family & \multicolumn{12}{l}{Full-rank Gaussian} \\
   & Step-search scheme & \multicolumn{12}{l}{Comprehensive step-search} \\
   & ${\nabla}_{\phi}$ & \multicolumn{6}{l}{Estimated without dropping the score-function term} & \multicolumn{6}{l}{Estimated with DReG} \\
   & LI & \multicolumn{3}{l}{Used} & \multicolumn{3}{l}{Not Used} & \multicolumn{3}{l}{Used} & \multicolumn{3}{l}{Not Used} \\
   & IWVI M$_{\text{training}}$ & \multicolumn{3}{l}{10} & \multicolumn{3}{l}{10} & \multicolumn{3}{l}{10} & \multicolumn{3}{l}{10} \\
   & IWVI M$_{\text{sampling}}$ & \multicolumn{3}{l}{10} & \multicolumn{3}{l}{10} & \multicolumn{3}{l}{10} & \multicolumn{3}{l}{10} \\
   & Method from Outline & \multicolumn{3}{l}{Additional} & \multicolumn{3}{l}{Additional} & \multicolumn{3}{l}{Additional} & \multicolumn{3}{l}{(3b)} \\
   & Independent Trial &                   Trial 1 &     Trial 2 &     Trial 3 &     Trial 1 &     Trial 2 &     Trial 3 &     Trial 1 &     Trial 2 &     Trial 3 &     Trial 1 &     Trial 2 &     Trial 3 \\
\bottomrule
\end{tabular}
}
\end{small}
\end{center}
\end{table}

\end{landscape}

\begin{landscape}
\thispagestyle{empty} 
\begin{table}[h!]
\caption{\label{tab:nf_vi} This table provides results for real-NVP normalizing flows optimized with our comprehensive step-search scheme with additional results from using IW-sampling.}
\begin{center}
\begin{small}
\resizebox{\linewidth}{!}
{
	\begin{tabular}{llrrrrrrrrrrrr}
	\toprule
	   & $q_{\phi}$ family & \multicolumn{12}{l}{Real NVP flows} \\
	   & Step-search scheme & \multicolumn{12}{l}{Comprehensive step-search} \\
	   & ${\nabla}_{\phi}$ & \multicolumn{6}{l}{Estimated without dropping the score-function term--full gradient} & \multicolumn{6}{l}{Estimated with STL} \\
	   & LI & \multicolumn{6}{l}{Not Used} & \multicolumn{6}{l}{Not Used} \\
	   & IWVI M$_{\text{training}}$ & \multicolumn{6}{l}{1} & \multicolumn{6}{l}{1} \\
	   & IWVI M$_{\text{sampling}}$ & \multicolumn{3}{l}{1} & \multicolumn{3}{l}{10} & \multicolumn{3}{l}{1} & \multicolumn{3}{l}{10} \\
	   & Method from Outline & \multicolumn{3}{l}{(4a)} & \multicolumn{3}{l}{Additional} & \multicolumn{3}{l}{(4b)} & \multicolumn{3}{l}{(4c)} \\
	   & Independent Trial &                   Trial 1 &     Trial 2 &     Trial 3 &     Trial 1 &     Trial 2 &     Trial 3 &     Trial 1 &     Trial 2 &     Trial 3 &     Trial 1 &     Trial 2 &     Trial 3 \\
\bottomrule
	\end{tabular}
}
\end{small}
\end{center}
\end{table}

\end{landscape}

\begin{landscape}
\thispagestyle{empty} 
\begin{table}[!ht]
\caption{\label{tab:nf_iwvi} This table provides results for importance weighted training for real-NVP normalizing flows optimized with our comprehensive step-search scheme with additional results from using regular IW-ELBO gradient.}
\begin{center}
\begin{small}
\resizebox{0.6\linewidth}{!}
{
	
	\begin{tabular}{llrrrrrr}
\toprule
   & $q_{\phi}$ family & \multicolumn{6}{l}{Real NVP flows} \\
   & Step-search scheme & \multicolumn{6}{l}{Comprehensive step-search} \\
   & ${\nabla}_{\phi}$ & \multicolumn{3}{l}{Estimated w/o dropping score-function term} & \multicolumn{3}{l}{Estimated with DReG} \\
   & LI & \multicolumn{3}{l}{Not Used} & \multicolumn{3}{l}{Not Used} \\
   & IWVI M$_{\text{training}}$ & \multicolumn{3}{l}{10} & \multicolumn{3}{l}{10} \\
   & IWVI M$_{\text{sampling}}$ & \multicolumn{3}{l}{10} & \multicolumn{3}{l}{10} \\
   & Method from Outline & \multicolumn{3}{l}{Additional} & \multicolumn{3}{l}{(4d)} \\
   & Independent Trial &                   Trial 1 &     Trial 2 &     Trial 3 &     Trial 1 &     Trial 2 &     Trial 3 \\
\bottomrule
\end{tabular}
}
\end{small}
\end{center}
\end{table}

\end{landscape}

\begin{landscape}
\thispagestyle{empty} 
\begin{table}[h!]
\vspace{-50pt}
\caption{This table presents the results for additional Diagonal Gaussian experiments. Please refer to \Cref{fig:diag-vs-fullrank for g,sec:extended results} for more details. }
\label{tab:diag-results}
\begin{center}
\begin{small}
\resizebox{0.8\linewidth}{!}
{
\begin{tabular}{llrrrrrrrrr}
\toprule
   & $q_{\phi}$ family & \multicolumn{9}{l}{Diagonal Gaussian} \\
   & Step-search scheme & \multicolumn{9}{l}{Comprehensive step-search} \\
   & ${\nabla}_{\phi}$ & \multicolumn{3}{l}{Closed form entropy} & \multicolumn{6}{l}{Estimated with STL} \\
   & LI & \multicolumn{3}{l}{Not Used} & \multicolumn{6}{l}{Not Used} \\
   & IWVI M$_{\text{training}}$ & \multicolumn{3}{l}{1} & \multicolumn{6}{l}{1} \\
   & IWVI M$_{\text{sampling}}$ & \multicolumn{3}{l}{1} & \multicolumn{3}{l}{1} & \multicolumn{3}{l}{10} \\
   & Method from Outline & \multicolumn{3}{l}{Additional} & \multicolumn{3}{l}{Additional} & \multicolumn{3}{l}{Additional} \\
   & Independent Trial &                   Trial 1 &     Trial 2 &     Trial 3 &     Trial 1 &     Trial 2 &     Trial 3 &     Trial 1 &     Trial 2 &     Trial 3 \\
\bottomrule
\end{tabular}

}
\end{small}
\end{center}
\end{table}

\end{landscape}

\section{Complete table for per iteration training times}
\label{sec:complete-time-tables}
For completeness, we include the per iterations training times of all the VI methods we experiment with. However, these training times should be read into with caution. We interface with Pystan and Autograd for our work; this creates an extra overhead with can dominate the run-times when the models are expensive to evaluate. Further, each training instance is run on a single CPU core.
\begin{landscape}
\thispagestyle{empty} 
\begin{table}[h!]
\vspace{-50pt}
\caption{This table presents the per iteration training times for ADVI baseline. Please refer to \Cref{tab:advi-unstable} for lower-bound results.}
\label{tab:advi-time}
\begin{center}
\begin{small}
\resizebox{0.35\linewidth}{!}
{
\begin{tabular}{llrrr}
\toprule
   & $q_{\phi}$ family & \multicolumn{3}{l}{Full-rank Gaussian} \\
   & Step-search scheme & \multicolumn{3}{l}{ADVI} \\
   & ${\nabla}_{\phi}$ & \multicolumn{3}{l}{Closed form entropy} \\
   & LI & \multicolumn{3}{l}{Not Used} \\
   & IWVI M$_{\text{training}}$ & \multicolumn{3}{l}{1} \\
   & IWVI M$_{\text{sampling}}$ & \multicolumn{3}{l}{1} \\
   & Independent Trial &             Trial 1 & Trial 2 & Trial 3 \\
Id & Model Name &                     &         &         \\
\bottomrule
\end{tabular}

}
\end{small}
\end{center}
\end{table}

\end{landscape}
\begin{landscape}
\thispagestyle{empty} 
\begin{table}[h!]
\caption{This table provides the per iteration training times for method that uses the closed-form entropy gradient with our comprehensive step-search scheme. Refer to \Cref{tab:g_closed_form} for lower-bound results.}
\label{tab:g_closed_form_time}
\begin{center}
\begin{small}
\resizebox{0.7\linewidth}{!}
{\begin{tabular}{llrrrrrrrrrrrr}
\toprule
   & $q_{\phi}$ family & \multicolumn{12}{l}{Full-rank Gaussian} \\
   & Step-search scheme & \multicolumn{12}{l}{Comprehensive step-search} \\
   & ${\nabla}_{\phi}$ & \multicolumn{12}{l}{Closed form entropy} \\
   & LI & \multicolumn{6}{l}{Not Used} & \multicolumn{6}{l}{Used} \\
   & IWVI M$_{\text{training}}$ & \multicolumn{6}{l}{1} & \multicolumn{6}{l}{1} \\
   & IWVI M$_{\text{sampling}}$ & \multicolumn{3}{l}{1} & \multicolumn{3}{l}{10} & \multicolumn{3}{l}{1} & \multicolumn{3}{l}{10} \\
   & Independent Trial &                   Trial 1 & Trial 2 & Trial 3 & Trial 1 & Trial 2 & Trial 3 & Trial 1 & Trial 2 & Trial 3 & Trial 1 & Trial 2 & Trial 3 \\
Id & Model Name &                           &         &         &         &         &         &         &         &         &         &         &         \\
\bottomrule
\end{tabular}
}
\end{small}
\end{center}
\end{table}

\end{landscape}
\begin{landscape}
\thispagestyle{empty} 
\begin{table}[h!]
\caption{\label{tab:g_wo_stl_time} This table provides per iteration training times for Gaussian VI that uses the ``full'' entropy gradient from \Cref{eq:entropy-gradients} with our comprehensive step-search scheme. Refer to \Cref{tab:g_wo_stl} for lower-bound results.}
\begin{center}
\begin{small}
\resizebox{0.8\linewidth}{!}
{\begin{tabular}{llrrrrrrrrrrrr}
\toprule
   & $q_{\phi}$ family & \multicolumn{12}{l}{Full-rank Gaussian} \\
   & Step-search scheme & \multicolumn{12}{l}{Comprehensive step-search} \\
   & ${\nabla}_{\phi}$ & \multicolumn{12}{l}{Estimated without dropping score function term--full gradient} \\
   & LI & \multicolumn{6}{l}{Used} & \multicolumn{6}{l}{Not Used} \\
   & IWVI M$_{\text{training}}$ & \multicolumn{6}{l}{1} & \multicolumn{6}{l}{1} \\
   & IWVI M$_{\text{sampling}}$ & \multicolumn{3}{l}{1} & \multicolumn{3}{l}{10} & \multicolumn{3}{l}{1} & \multicolumn{3}{l}{10} \\
   & Independent Trial &                   Trial 1 & Trial 2 & Trial 3 & Trial 1 & Trial 2 & Trial 3 &  Trial 1 & Trial 2 & Trial 3 & Trial 1 & Trial 2 & Trial 3 \\
Id & Model Name &                           &         &         &         &         &         &          &         &         &         &         &         \\
\bottomrule
\end{tabular}
}
\end{small}
\end{center}
\end{table}

\end{landscape}
\begin{landscape}
\thispagestyle{empty} 
\begin{table}[h!]
\caption{\label{tab:g_w_stl_time}  This table provides per iteration training times when using STL gradient from \Cref{eq:entropy-gradients} with our comprehensive step-search scheme. Please refer to \Cref{tab:g_w_stl} for lower-bound results.}
\begin{center}
\begin{small}
\resizebox{0.8\linewidth}{!}
{\begin{tabular}{llrrrrrrrrrrrr}
\toprule
   & $q_{\phi}$ family & \multicolumn{12}{l}{Full-rank Gaussian} \\
   & Step-search scheme & \multicolumn{12}{l}{Comprehensive step-search} \\
   & ${\nabla}_{\phi}$ & \multicolumn{12}{l}{Estimate with STL} \\
   & LI & \multicolumn{6}{l}{Used} & \multicolumn{6}{l}{Not Used} \\
   & IWVI M$_{\text{training}}$ & \multicolumn{6}{l}{1} & \multicolumn{6}{l}{1} \\
   & IWVI M$_{\text{sampling}}$ & \multicolumn{3}{l}{1} & \multicolumn{3}{l}{10} & \multicolumn{3}{l}{1} & \multicolumn{3}{l}{10} \\
   & Independent Trial &                   Trial 1 & Trial 2 & Trial 3 & Trial 1 & Trial 2 & Trial 3 &  Trial 1 & Trial 2 & Trial 3 & Trial 1 & Trial 2 & Trial 3 \\
Id & Model Name &                           &         &         &         &         &         &          &         &         &         &         &         \\
\bottomrule
\end{tabular}}
\end{small}
\end{center}
\end{table}

\end{landscape}
\begin{landscape}
\thispagestyle{empty} 
\begin{table}[h!]
\caption{\label{tab:g_iwvi_time} This table provides per iteration training times for importance-weighted training for Gaussian $q_{\phi}$ optimized with our comprehensive step-search scheme. Please refer to \Cref{tab:g_iwvi} for lower-bound results.}
\begin{center}
\begin{small}
\resizebox{0.8\linewidth}{!}
{\begin{tabular}{llrrrrrrrrrrrr}
\toprule
   & $q_{\phi}$ family & \multicolumn{12}{l}{Full-rank Gaussian} \\
   & Step-search scheme & \multicolumn{12}{l}{Comprehensive step-search} \\
   & ${\nabla}_{\phi}$ & \multicolumn{6}{l}{Estimated without dropping the score-function term} & \multicolumn{6}{l}{Estimated with DReG} \\
   & LI & \multicolumn{3}{l}{Used} & \multicolumn{3}{l}{Not Used} & \multicolumn{3}{l}{Used} & \multicolumn{3}{l}{Not Used} \\
   & IWVI M$_{\text{training}}$ & \multicolumn{3}{l}{10} & \multicolumn{3}{l}{10} & \multicolumn{3}{l}{10} & \multicolumn{3}{l}{10} \\
   & IWVI M$_{\text{sampling}}$ & \multicolumn{3}{l}{10} & \multicolumn{3}{l}{10} & \multicolumn{3}{l}{10} & \multicolumn{3}{l}{10} \\
   & Independent Trial &                   Trial 1 & Trial 2 & Trial 3 &  Trial 1 & Trial 2 & Trial 3 &  Trial 1 & Trial 2 & Trial 3 &  Trial 1 & Trial 2 & Trial 3 \\
Id & Model Name &                           &         &         &          &         &         &          &         &         &          &         &         \\
\bottomrule
\end{tabular}
}
\end{small}
\end{center}
\end{table}

\end{landscape}
\begin{landscape}
\thispagestyle{empty} 
\begin{table}[h!]
\caption{\label{tab:nf_vi_time} This table provides per iterations training times for real-NVP normalizing flows optimized with our comprehensive step-search scheme with additional results from using IW-sampling. Please refer to \Cref{tab:nf_vi} for lower-bound results.}
\begin{center}
\begin{small}
\resizebox{0.7\linewidth}{!}
{\begin{tabular}{llrrrrrrrrrrrr}
\toprule
   & $q_{\phi}$ family & \multicolumn{12}{l}{Real NVP flows} \\
   & Step-search scheme & \multicolumn{12}{l}{Comprehensive step-search} \\
   & ${\nabla}_{\phi}$ & \multicolumn{6}{l}{Estimated without dropping the score-function term} & \multicolumn{6}{l}{Estimated with STL} \\
   & & \multicolumn{6}{l}{--full gradient} & \multicolumn{6}{l}{} \\
   & LI & \multicolumn{6}{l}{Not Used} & \multicolumn{6}{l}{Not Used} \\
   & IWVI M$_{\text{training}}$ & \multicolumn{6}{l}{1} & \multicolumn{6}{l}{1} \\
   & IWVI M$_{\text{sampling}}$ & \multicolumn{3}{l}{1} & \multicolumn{3}{l}{10} & \multicolumn{3}{l}{1} & \multicolumn{3}{l}{10} \\
   & Independent Trial &                   Trial 1 & Trial 2 & Trial 3 & Trial 1 & Trial 2 & Trial 3 &  Trial 1 & Trial 2 & Trial 3 & Trial 1 & Trial 2 & Trial 3 \\
\bottomrule
\end{tabular}
}
\end{small}
\end{center}
\end{table}

\end{landscape}
\begin{landscape}
\thispagestyle{empty} 
\begin{table}[!ht]
\caption{\label{tab:nf_iwvi_time} This table provides per iteration training times for importance weighted training for real-NVP normalizing flows optimized with our comprehensive step-search scheme with additional results from using regular IW-ELBO gradient. Please refer to \Cref{tab:nf_iwvi} for lower-bound results.}
\begin{center}
\begin{small}
\resizebox{0.4\linewidth}{!}
{\begin{tabular}{llrrrrrr}
\toprule
   & $q_{\phi}$ family & \multicolumn{6}{l}{Real NVP flows} \\
   & Step-search scheme & \multicolumn{6}{l}{Comprehensive step-search} \\
   & ${\nabla}_{\phi}$ & \multicolumn{3}{l}{Estimated w/o dropping } & \multicolumn{3}{l}{Estimated with DReG} \\
   & & \multicolumn{3}{l}{score-function term} & \multicolumn{3}{l}{} \\
   & LI & \multicolumn{3}{l}{Not Used} & \multicolumn{3}{l}{Not Used} \\
   & IWVI M$_{\text{training}}$ & \multicolumn{3}{l}{10} & \multicolumn{3}{l}{10} \\
   & IWVI M$_{\text{sampling}}$ & \multicolumn{3}{l}{10} & \multicolumn{3}{l}{10} \\
   & Independent Trial &                   Trial 1 & Trial 2 & Trial 3 &  Trial 1 & Trial 2 & Trial 3 \\
\bottomrule
\end{tabular}}
\end{small}
\end{center}
\end{table}

\end{landscape}
\begin{landscape}
\thispagestyle{empty} 
\begin{table}[h!]
\vspace{-50pt}
\caption{This table presents the per iteration training time for additional Diagonal Gaussian experiments. Please refer to \Cref{fig:diag-vs-fullrank for g,sec:extended results} for more details. }
\label{tab:diag-time}
\begin{center}
\begin{small}
\resizebox{0.6\linewidth}{!}
{\begin{tabular}{llrrrrrrrrr}
\toprule
   & $q_{\phi}$ family & \multicolumn{9}{l}{Diagonal Gaussian} \\
   & Step-search scheme & \multicolumn{9}{l}{Comprehensive step-search} \\
   & ${\nabla}_{\phi}$ & \multicolumn{3}{l}{Closed form entropy} & \multicolumn{6}{l}{Estimated with STL} \\
   & LI & \multicolumn{3}{l}{Not Used} & \multicolumn{6}{l}{Not Used} \\
   & IWVI M$_{\text{training}}$ & \multicolumn{3}{l}{1} & \multicolumn{6}{l}{1} \\
   & IWVI M$_{\text{sampling}}$ & \multicolumn{3}{l}{1} & \multicolumn{3}{l}{1} & \multicolumn{3}{l}{10} \\
   & Independent Trial &                   Trial 1 & Trial 2 & Trial 3 &  Trial 1 & Trial 2 & Trial 3 & Trial 1 & Trial 2 & Trial 3 \\
Id & Model Name &                           &         &         &          &         &         &         &         &         \\
\bottomrule
\end{tabular}
}
\end{small}
\end{center}
\end{table}

\end{landscape}

\end{document}